\documentclass[preprint,12pt,authoryear]{elsarticle}

\usepackage{amssymb}
\usepackage{booktabs}
\usepackage{graphicx}
\usepackage{hyperref}
\usepackage{placeins}
\usepackage[utf8]{inputenc}

\usepackage[toc]{appendix}
\usepackage{svg}
\usepackage{algorithm,algpseudocode}
\usepackage[utf8]{inputenc}
\usepackage{amssymb}
\usepackage{pifont} 
\newcommand{\cmark}{\ding{51}} 
\newcommand{\xmark}{\ding{55}} 
\DeclareUnicodeCharacter{03BA}{\ensuremath{\kappa}}
\usepackage{amsmath}
\usepackage{doi}
\usepackage{textcomp} 
\usepackage{csquotes}
\usepackage[utf8]{inputenc}
\DeclareUnicodeCharacter{2013}{--}
\usepackage[T1]{fontenc}

\journal{Nuclear Physics B}

\begin{document}

\begin{frontmatter}



\title{LoLA-SpecViT: Local Attention SwiGLU Vision Transformer with LoRA for Hyperspectral Imaging
} 


\author{Fadi Abdeladhim Zidi} 
\affiliation{%
  organization={VSC Laboratory, Department of Electrical Engineering}, 
  addressline={University of Mohamed Khider}, 
  city={Biskra},
  postcode={07000}, 
  state={Biskra},
  country={Algeria}
}

\author{Djamel Eddine Boukhari} 
\affiliation{%
  organization={Scientific and Technical Research Centre for Arid Areas (CRSTRA)}, 
  city={Biskra},
  postcode={07000}, 
  state={Biskra},
  country={Algeria}
}

\author{Abdellah Zakaria Sellam} 
\affiliation{%
  organization={Department of Innovation Engineering, University of Salento \& Institute of Applied Sciences and Intelligent Systems – CNR}, 
  addressline={Via per Monteroni}, 
  city={Lecce},
  postcode={73100}, 
  state={Lecce},
  country={Italy}
}

\author{Abdelkrim Ouafi} 
\affiliation{%
  organization={VSC Laboratory, Department of Electrical Engineering}, 
  addressline={University of Mohamed Khider}, 
  city={Biskra},
  postcode={07000}, 
  state={Biskra},
  country={Algeria}
}

\author{Cosimo Distante} 
\affiliation{%
  organization={Institute of Applied Sciences and Intelligent Systems – CNR}, 
  addressline={Via per Monteroni}, 
  city={Lecce},
  postcode={73100}, 
  state={Lecce},
  country={Italy}
}
\author{Salah Eddine Bekhouche} 
\affiliation{%
  organization={UPV/EHU}, 
  addressline={University of the Basque Country}, 
  city={Sebastian},
  postcode={20018}, 
  state={Sebastian},
  country={Spain}
}

\author{Abdelmalik Taleb-Ahmed} 
\affiliation{%
  organization={Université Polytechnique Hauts-de-France, Université de Lille, CNRS}, 
  addressline={}, 
  city={Valenciennes},
  postcode={59313}, 
  state={Hauts-de-France},
  country={France}
}

\begin{abstract}
Hyperspectral image classification remains a challenging task due to the high dimensionality of spectral data, significant inter-band redundancy, and the limited availability of annotated samples. While recent transformer-based models have improved the global modeling of spectral-spatial dependencies, their scalability and adaptability under label-scarce conditions remain limited. In this work, we propose \textbf{LoLA-SpecViT}(Low-rank adaptation Local Attention Spectral Vision Transformer), a lightweight spectral vision transformer that addresses these limitations through a parameter-efficient architecture tailored to the unique characteristics of hyperspectral imagery.
Our model combines a 3D convolutional spectral front-end with local window-based self-attention, enhancing both spectral feature extraction and spatial consistency while reducing computational complexity. To further improve adaptability, we integrate low-rank adaptation (LoRA) into attention and projection layers, enabling fine-tuning with over 80\% fewer trainable parameters. A novel cyclical learning rate scheduler modulates LoRA adaptation strength during training, improving convergence and generalisation.
Extensive experiments on three benchmark datasets WHU-Hi LongKou, WHU-Hi HongHu, and Salinas demonstrate that LoLA-SpecViT consistently outperforms state-of-the-art baselines, achieving up to 99.91\% accuracy with substantially fewer parameters and enhanced robustness under low-label regimes. The proposed framework provides a scalable and generalizable solution for real-world HSI applications in agriculture, environmental monitoring, and remote sensing analytics. Our code is available in the following \href{https://github.com/FadiZidiDz/LoLA-SpecViT}{GitHub Repository}.
\end{abstract}

\begin{graphicalabstract}
\end{graphicalabstract}

\begin{highlights}
\item LoLA‑SpecViT: hierarchical spectral Vision Transformer with local windowed attention
\item LoRA fine‑tuning reduces trainable parameters by over 80\% for efficiency
\item SwiGLU activations and cyclical LoRA scheduler boost convergence and generalisation
\item Achieves up to 99.91\% overall accuracy on WHU‑LongKou, HongHu and Salinas datasets
\end{highlights}

\begin{keyword}
Hyperspectral image classification, Vision Transformer, Parameter-efficient fine-tuning, Low-rank adaptation, Spectral attention, Local self-attention.



\end{keyword}

\end{frontmatter}


\section{Introduction}
Hyperspectral imaging (HSI) has fundamentally transformed observational remote sensing through its capacity to capture hundreds of contiguous, narrow spectral bands at each ground location, generating comprehensive three-dimensional data cubes that substantially exceed the limited information provided by RGB or multispectral sensors \cite{zhang2021difference, huang2024spectral}. This enhanced spectral resolution has revolutionized agricultural monitoring capabilities by delivering unprecedented precision in crop analysis \citep{zhou2023precisionHSI}. Within agricultural contexts, these spectral signatures reveal subtle biochemical and structural characteristics that include early stress indicators, nutrient deficiencies, and moisture variations, all of which remain undetectable to conventional imaging systems. For instance, hyperspectral data enables early detection of drought stress or pathogen infection before visible symptoms appear, facilitates precise discrimination between crops and weeds for targeted herbicide application, and supports accurate estimation of chlorophyll levels and crop maturity for optimal harvest timing \cite{garcia2024hyperspectral, abdelkrim2024hyperspectral, wang2024characterization, zidi2025advancing}. These capabilities highlight why precision agriculture has emerged as a key application domain for HSI technology \cite{ram2024systematic}. However, these significant opportunities are accompanied by substantial technical challenges that hinder HSI's practical implementation. The curse of dimensionality frequently overwhelms classical classification algorithms, while spectral variability from illumination and atmospheric conditions can severely compromise material identification accuracy. Furthermore, the fundamental trade-off between spectral richness and spatial resolution often limits effective real-world deployment \cite{plaza2019dimensionality, chen2020variability, li2022trades}. These concurrent opportunities and obstacles necessitate the development of innovative deep-learning frameworks capable of efficiently distilling high-dimensional spectra into robust, discriminative features while preserving critical spatial-spectral relationships \cite{ma2023deep}.

Early deep-learning methods for HSI classification focused primarily on convolutional architectures. Li et al.\cite{li2017spectral} introduced a 3D‑CNN to jointly learn local spectral-spatial features, demonstrating strong performance on datasets with limited labeled samples. Subsequent studies confirmed that while CNNs effectively extract hierarchical features, they require millions of parameters and extensive data augmentation to capture long-range dependencies \cite{liu2025receptiveField}.
In response, transformer-based architectures have emerged as a compelling alternative. By tokenising hyperspectral cubes and applying self-attention, Vision Transformers (ViTs) can model global spectral-spatial dependencies without relying on deep convolutional stacks \citep{he2024hsiViT, chen2024surveyTransformerHSI}. However, the quadratic complexity of standard self-attention concerning sequence length makes vanilla ViTs computationally prohibitive for large-scale HSI volumes.
To reconcile locality and global context, recent hybrid CNN–Transformer architectures interleave multi-scale convolutions with windowed self-attention. For instance, Cai et al.\citep{cai2022coarseFineSparse} proposed a coarse‑to‑fine sparse attention mechanism, while Hu et al.\citep{hu2024selectiveTransformerHSI} introduced selective spectral attention windows to prune redundant computations. Despite these advances, hybrid designs often underutilise inter-band correlations or retain large parameter footprints—challenges directly addressed by our proposed LoLA‑SpecViT architecture.

To systematically address these identified bottlenecks while leveraging the proven strengths of transformer architectures, we introduce LoLA-SpecViT, a hierarchical vision transformer architecture that strategically combines a lightweight 3D convolutional spectral front-end with windowed self-attention blocks utilizing SwiGLU activations. Our approach addresses the computational limitations of standard transformers while preserving their global modeling capabilities. We further enhance both efficiency and generalisation through the implementation of a novel BandDropout mechanism for spectral regularisation and a specialised spectral attention module that explicitly models inter-band correlations. Crucially, we employ Parameter-Efficient Fine-Tuning (PEFT) through a cyclic LoRA rate schedule \citep{hu2021lora, lester2021power}, achieving a reduction in trainable parameters exceeding 80\% while preserving global contextual information and attaining state-of-the-art classification accuracy across benchmark HSI datasets. This design represents a strategic integration of transformer efficiency improvements with domain-specific innovations for agricultural hyperspectral imaging applications. The main contributions of this work are summarised as follows.
\begin{itemize}
    \item We propose LoLA-SpecViT, a novel hierarchical vision transformer architecture that combines windowed attention mechanisms with SwiGLU activations to model multi-scale spectral-spatial dependencies in hyperspectral data effectively.
    \item A spectral processing front-end based on 3D convolutions is designed with a novel BandDropout technique and spectral attention module to enhance band-wise feature learning while reducing overfitting risks.
    \item We incorporate parameter-efficient fine-tuning (PEFT) techniques, including a newly introduced Cyclic LoRA Rate (CLR) scheduling strategy, which improves training efficiency and significantly reduces computational costs.
    \item Extensive experiments on multiple benchmark hyperspectral datasets demonstrate that LoLA-SpecViT achieves state-of-the-art classification accuracy with lower resource requirements, confirming its effectiveness and superiority over existing methods.
\end{itemize}
The remainder of this paper is structured as follows: Section~\ref{sec:related_work} surveys the relevant literature; Section~\ref{sec:methodology} details the LoLA‑SpecViT architecture; Section~\ref{sec:experiments} describes the experimental setup; Section~\ref{sec:results} presents and analyses the findings; and Section~\ref{sec:conclusion} concludes the study with suggestions for future research.

\section{Related Work}
\label{sec:related_work}
The field of HSI classification has progressed significantly over the past decade, shifting from conventional machine learning algorithms toward deep learning-based architectures designed to address the challenges of high-dimensional spectral-spatial analysis. This transition is shaped by the imperative to balance model expressiveness with scalability, particularly for applications requiring accurate, real-time interpretation of large-scale HSI data.
\subsection{\textbf{Deep Learning Foundations for HSI Classification}}
The transition from handcrafted to learned features has been transformative in HSI classification, enabling models to directly capture both spectral and spatial regularities from the data \citep{Chen2021_Survey}. Hu et al.\ \citep{Hu2015} initiated this shift with pixel‑wise multilayer perceptrons (MLPs), demonstrating baseline spectral discrimination but neglecting contextual information. Li et al.\ \citep{Li2017} subsequently introduced 2D convolutional neural networks (CNNs) for localized spatial filtering and later extended to 3D CNNs for unified spectral-spatial fusion, though at a significant computational expense. HybridSN \citep{Roy2020} elegantly balances this trade‑off by cascading 3D and 2D convolutions, yielding rich joint representations with controlled model complexity.
To further enhance feature adaptability, Yang et al.\ \citep{Yang2021A2MFE} proposed A²MFE, which automatically adjusts receptive fields via multi‑scale convolutions, and Sun et al.\ \citep{Sun2022} developed SSFTT, a spectral–spatial feature tokenization transformer that first extracts low‑level features through 3D–2D CNNs before feeding Gaussian‑weighted tokens into a Transformer encoder. Attention‑driven designs such as MADANet \cite{cui2023madanet}, which integrates multiscale feature aggregation with dual attention, and HiT \citep{Yang2022a}, which embeds spectral-adaptive 3D convolutions within a pure Transformer pipeline, have delivered further gains. Complementing these, LANet \citep{Ding2021} embeds local attention modules into a shallow encoder-decoder, enriching low‑level semantic features for improved segmentation and classification performance.
As these architectures deepen, their resource demands escalate, challenging deployment on edge and embedded platforms. Lei et al.\ \citep{Lei2023_CollaborativePruning} address redundancy via collaborative pruning, while Yue et al.\ \citep{Yue2022_SelfSupDistill} and Yu et al.\ \citep{Yu2024_PrototypeDistill} demonstrate that self-supervised and prototype-based distillation preserve accuracy under severe compression. More recently, Ma et al.\ \citep{Ma2024_AS2MLP} and Zhang et al.\ \citep{Zhang2024_LDS2MLP} introduced AS2MLP and LDS2MLP, respectively adaptive MLP variants that fuse convolutional priors with learnable spatial shifts, offering near-global receptive fields with minimal overhead.
Despite these advances, purely convolutional and shallow MLP frameworks remain challenged by the need to model long-range spectral dependencies across hundreds of contiguous bands, motivating the exploration of transformer‑based and hybrid paradigms.
\subsection{\textbf{Vision Transformers and Hybrid Architectures for Global Modeling}}
Vision Transformers (ViTs) reformulate images as sequences of tokens and apply self-attention to capture global dependencies, dispensing with convolutional locality constraints. Dosovitskiy et al.\ \citep{Dosovitskiy2021} first demonstrated this on RGB imagery; Ibáñez et al.\ \citep{Ibanez2022} then introduced HSIMAE, a masked autoencoder that pretrains robust spectral embeddings under label scarcity. Cross SSL \citep{bai2024cross} extends self‑supervision to learn representations transferable across sensors and environments.
Sun et al.\ \citep{Sun2024MASSFormer} further mitigate spectral redundancy with MASSFormer’s memory‑augmented self‑attention and spatial-spectral positional encoding. To control Transformer complexity, Zhou et al.\ \citep{Zhou2023} proposed DiCT, dynamically pruning low‑importance tokens.
Hybrid designs fuse local inductive biases with global attention. Arshad et al.\ \citep{Arshad2024} present Hybrid‑ViT, pairing a residual 3D CNN with a transformer and channel‑attention module to capture complementary cues. Zhang et al.\ \citep{Zhang2023}’s 3D‑CmT fuses 3D–CNN frontends with a transformer encoder for multiscale spectral-spatial modeling. Xu et al.\ \citep{Xu2024DBCT}’s DBCTNet employs dual branches one 3D CNN, and one Transformer for balanced context integration, while Ma et al.\ \citep{Ma2023LSGA}’s LSGA introduces light self‑Gaussian attention for efficient feature discrimination.
Recognizing the need for hierarchical abstraction, Ahmad et al.\ \citep{Ahmad2024} proposed the Hir‑Transformer (Pyramid Hierarchical Transformer), which organises tokens across multiple levels to capture both local detail and global structure effectively.
State‑space transformer variants now offer linear scalability. SSMamba and MorpMamba \citep{ahmad2025spatial} integrate structured state‑space modules with morphological operations, achieving long‑sequence modelling at linear cost. Finally, E‑SR‑SSIM \citep{hu2023improved} employs subspace partitioning guided by structural similarity to select representative bands, reducing redundancy while enhancing interpretability.
While these approaches successfully address the locality limitations of CNNs, they introduce their own computational challenges. The quadratic complexity of self-attention mechanisms concerning sequence length creates scalability issues when processing large HSI volumes, necessitating further innovations in efficient attention computation and model compression strategies.

\subsection{\textbf{Efficiency-Oriented Architectures and Optimization Strategies}}
Recognising the computational and memory constraints imposed by both deep CNNs and standard transformers, recent research has focused on developing lightweight architectures and optimisation strategies that maintain classification performance while reducing computational overhead. Khan et al.\ \citep{Khan2024} introduced GroupFormer, which partitions spectral bands into discrete groups and performs intra-group self-attention, effectively reducing computational complexity from quadratic to linear concerning the number of spectral bands without sacrificing classification accuracy.
Advancing this efficiency-focused paradigm, Zhu et al.\ \citep{Zhu2023} developed PatchOut, which replaces fixed-size spatial patching with learned token sampling mechanisms to streamline the encoding process. While this approach significantly reduces computational requirements, it potentially risks losing fine-grained spatial details crucial for accurate classification. Fu et al.\ \citep{Fu2025} incorporated Convolutional Block Attention Module (CBAM) based adaptive pruning to dynamically eliminate redundant channels and spatial locations during inference, providing runtime efficiency improvements.
These efficiency-oriented strategies achieve substantial reductions in parameter counts and inference times but often overlook cross-group dependencies in grouped approaches or require dataset-specific threshold tuning for pruning mechanisms. Furthermore, these methods typically focus on architectural modifications rather than addressing the fundamental challenge of adapting large-scale pre-trained models to the specific characteristics of hyperspectral data.
\subsection{\textbf{Efficient Adaptation of Large Pre-trained Models}}
The limitations identified in existing approaches highlight a critical gap in HSI classification research: the need for methods that can leverage the representational power of large pre-trained models while maintaining computational efficiency and adapting effectively to the unique characteristics of hyperspectral data. PEFT addresses this challenge by adapting large pre-trained models to downstream tasks through updating only a small fraction of parameters ($|\Delta\theta|\ll|\theta|$), thereby substantially reducing both computational and storage overheads \citep{Xin2024}.
PEFT approaches encompass several distinct methodologies. Adapter-based methods introduce lightweight trainable modules into each layer, exemplified by adapter tuning \citep{Houlsby2019} and visual prompt tuning (VPT) \citep{Jia2022}. Unified tuning approaches, such as NOAH, combine multiple adapter variants within a single framework to jointly optimize diverse modules for improved flexibility and performance \citep{Zhang2024_NOAH}. Partial-parameter update strategies restrict fine-tuning to selected parameter subsets, including Bias-Only tuning (BitFit) \citep{BenZaken2022}, low-rank adaptation via LoRA \citep{Hu2022}, its Kronecker-product extension KronA \citep{Edalati2022}, and the hybrid Lokr method \citep{Yeh2024}. Recent enhancements in LoRA\(^+\) demonstrate that decoupling learning rates between low-rank factors can further increase fine-tuning efficiency \citep{Hayou2024}.
\subsection{\textbf{Toward Generalizable and Efficient HSI Classification}}
The comprehensive analysis of existing literature reveals several persistent challenges in HSI classification. Convolutional approaches, while computationally efficient, remain constrained by local receptive fields and struggle to capture long-range spectral dependencies. Standard transformer architectures successfully model global relationships but suffer from quadratic computational complexity. Efficiency-focused methods reduce computational overhead but often sacrifice either accuracy or cross-dependency modeling capabilities. Most critically, existing approaches fail to effectively leverage semantic knowledge from large-scale pre-trained models that could provide valuable contextual understanding for spectral-spatial feature interpretation.
\FloatBarrier
\section{Proposed Method}
\label{sec:methodology}
\begin{figure}[htbp]
    \centering
    \includegraphics[width=15cm]{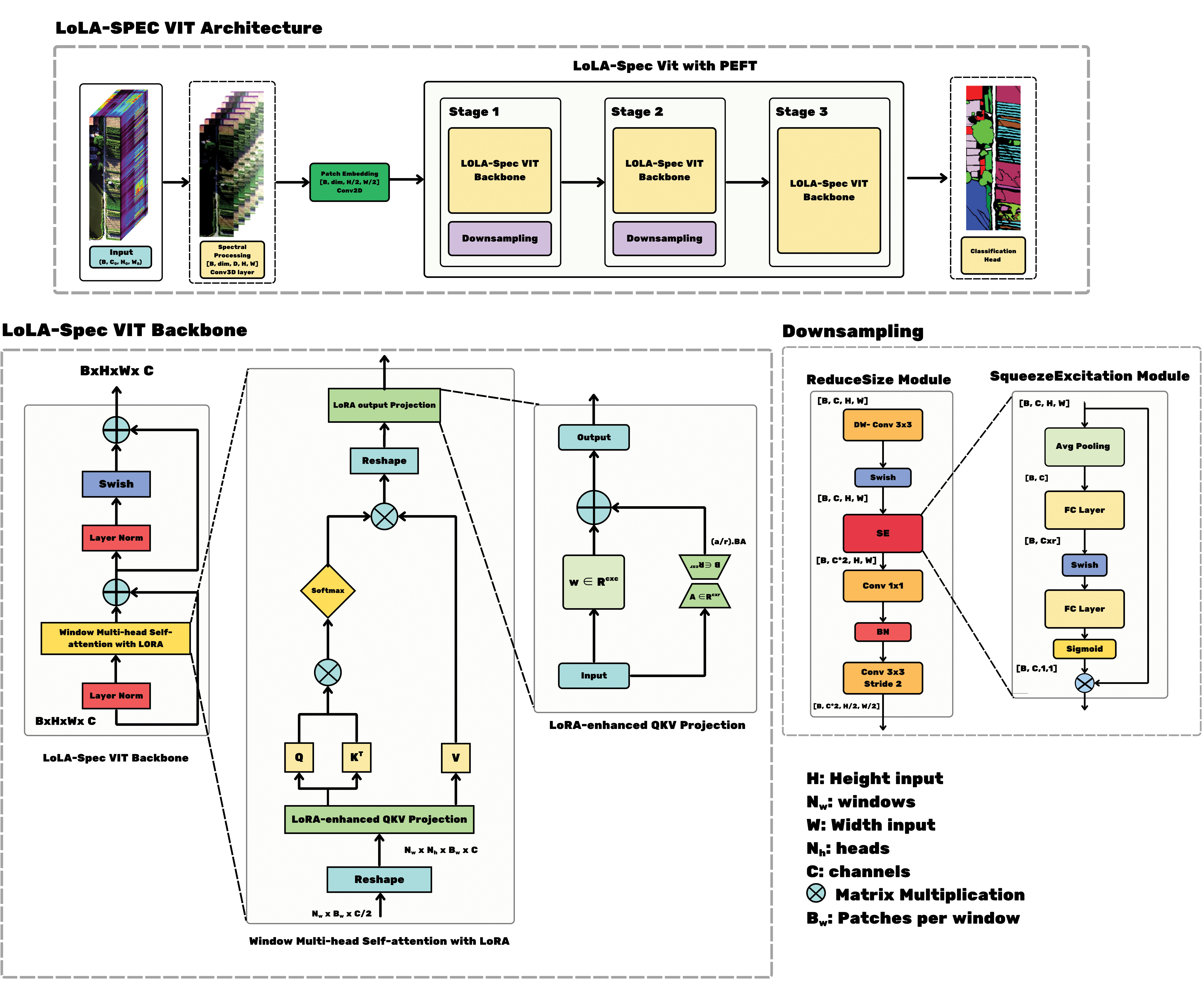}
    \caption{Architecture of the proposed LoLA-SpecViT for hyperspectral image classification.}
    \label{fig:Architecture}
\end{figure}

This section presents the proposed \textbf{LOLA-SpecViT} (Low-Rank Adapted Vision Transformer with Spectral Optimization), a novel parameter-efficient architecture tailored for hyperspectral image classification. The architecture, illustrated in Figure \ref{fig:Architecture}, integrates spectral-spatial feature learning with parameter-efficient fine-tuning through Low-Rank Adaptation (LoRA). LOLA-SpecViT comprises four main components: (1) PCA-based dimensionality reduction, (2) spectral processing front-end, (3) hierarchical transformer backbone with LoRA integration, and (4) classification head with enhanced feature projection.
\subsection{\textbf{Input Preprocessing and PCA Dimensionality Reduction}}
Given a raw hyperspectral image $\mathbf{X}_{\text{raw}} \in \mathbb{R}^{H \times W \times C_{\text{raw}}}$ where $H \times W$ represents spatial dimensions and $C_{\text{raw}}$ is the original number of spectral bands, we first apply Principal Component Analysis (PCA) for dimensionality reduction and spectral decorrelation.
The input is reshaped to matrix form and centred:
\begin{equation}
\mathbf{X}_{\text{flat}} = \text{Reshape}(\mathbf{X}_{\text{raw}}) \in \mathbb{R}^{(H \times W) \times C_{\text{raw}}}
\end{equation}
\begin{equation}
\mathbf{\Sigma} = \frac{1}{HW-1}(\mathbf{X}_{\text{flat}} - \boldsymbol{\mu})^T(\mathbf{X}_{\text{flat}} - \boldsymbol{\mu})
\end{equation}
where $\boldsymbol{\mu}$ is the mean spectral vector.
The PCA transformation with whitening is computed as follows:
\begin{equation}
\mathbf{\Sigma} = \mathbf{V}\mathbf{\Lambda}\mathbf{V}^T
\end{equation}
\begin{equation}
\mathbf{X}_{\text{pca}} = (\mathbf{X}_{\text{flat}} - \boldsymbol{\mu})\mathbf{V}_{k}\mathbf{\Lambda}_{k}^{-1/2}
\end{equation}
where $\mathbf{V}_{k} \in \mathbb{R}^{C_{\text{raw}} \times k}$ contains the first $k$ eigenvectors corresponding to the largest eigenvalues, and $\mathbf{\Lambda}_{k}$ contains the corresponding eigenvalues. The result is reshaped back to spatial format:
\begin{equation}
\mathbf{X}_{\text{in}} = \text{Reshape}(\mathbf{X}_{\text{pca}}) \in \mathbb{R}^{H \times W \times k}
\end{equation}
\subsubsection{\textbf{Patch Extraction and Tensor Preparation}}
For network processing, spatial patches of size $p \times p$ are extracted and prepared for batch processing:
\begin{equation}
F_{\text{in}} = \text{BatchPatch}(\mathbf{X}_{\text{in}}) \in \mathbb{R}^{B \times k \times p \times p}
\end{equation}
where $B$ is the batch size, $k$ is the number of principal components, and $p$ is the spatial patch size.
\subsection{\textbf{Spectral Processing Front-End}}
The spectral processing stage extracts hierarchical spectral-spatial features through a series of 3D convolutions. The input tensor is first reshaped to enable 3D convolution operations:
\begin{equation}
\tilde{F}_{\text{in}} = \text{Unsqueeze}(F_{\text{in}}) \in \mathbb{R}^{B \times 1 \times k \times p \times p}
\end{equation}

The spectral feature extraction follows a progressive convolution pipeline:
\begin{align}
F_1 &= \text{Swish}\left( \text{BatchNorm3d} \left( \text{Conv3d}_{(7,3,3)}(\tilde{F}_{\text{in}}) \right) \right) \label{eq:spectral_conv1} \\
F_2 &= \text{Swish}\left( \text{BatchNorm3d} \left( \text{Conv3d}_{(5,3,3)}(F_1) \right) \right) \label{eq:spectral_conv2} \\
F_{\text{spec}} &= \text{Swish}\left( \text{BatchNorm3d} \left( \text{Conv3d}_{(3,3,3)}(F_2) \right) \right) \label{eq:spectral_conv3}
,\end{align}
where $\text{Conv3d}_{(k_d, k_h, k_w)}$ denotes 3D convolution with kernel size $(k_d, k_h, k_w)$ along the depth (spectral), height, and width dimensions, respectively. The resulting output tensor $F_{\text{spec}}$ takes the form $\mathbb{R}^{B \times \text{dim} \times D \times H \times W}$, where $D$ denotes the spectral dimension after the 3D convolutions, corresponding to the number of processed spectral features. The Swish activation function is defined as:
\begin{equation}
\text{Swish}(x) = x \cdot \sigma(x) = x \cdot \frac{1}{1 + e^{-x}}
\end{equation}
This activation function promotes better gradient flow compared to ReLU and has been shown to improve convergence in deep networks.
\subsubsection{\textbf{Band Dropout Regularization}}
To enhance spectral robustness and prevent overfitting to specific spectral patterns, we introduce a \textbf{BandDropout} regularisation mechanism that randomly masks spectral channels during training:
\begin{equation}
M \sim \text{Bernoulli}(1 - p_{\text{drop}})
\end{equation}
\begin{equation}
F_{\text{bdrop}} = \frac{F_{\text{spec}} \odot M}{1 - p_{\text{drop}}}
\end{equation}
where $M \in \{0,1\}^{C_{\text{spec}}}$ is a binary mask sampled from a Bernoulli distribution with dropout probability $p_{\text{drop}}$, $\odot$ denotes element-wise multiplication, and the normalization factor $(1 - p_{\text{drop}})^{-1}$ maintains the expected magnitude of activations.
\subsubsection{\textbf{Spectral Attention Mechanism}}
To adaptively emphasize the most discriminative spectral features, we employ a \textbf{SpectralAttention} module that learns channel-wise attention weights:
\textbf{Global Spectral Pooling:}
\begin{equation}
\mathbf{z} = \text{GAP}(F_{\text{bdrop}}) \in \mathbb{R}^{B \times C_{\text{spec}}}
\end{equation}
where GAP denotes Global Average Pooling over spatial dimensions:
\begin{equation}
z_c = \frac{1}{H_{\text{spec}} \cdot W_{\text{spec}} \cdot D_{\text{spec}}} \sum_{d=1}^{D_{\text{spec}}} \sum_{h=1}^{H_{\text{spec}}} \sum_{w=1}^{W_{\text{spec}}} F_{\text{bdrop}}(b, c, d, h, w)
\end{equation}

The attention weights are computed through a two-layer MLP with bottleneck architecture:
\begin{equation}
A_{\text{spec}} = \sigma \left( \mathbf{W}_2 \cdot \text{Swish}( \mathbf{W}_1 \cdot \mathbf{z} ) \right)
\end{equation}
where $\mathbf{W}_1 \in \mathbb{R}^{r \times C_{\text{spec}}}$ and $\mathbf{W}_2 \in \mathbb{R}^{C_{\text{spec}} \times r}$ are learned projection matrices, $r < C_{\text{spec}}$ is the bottleneck dimension (typically $r = C_{\text{spec}}/4$), and $\sigma(\cdot)$ is the sigmoid activation function.
\textbf{Feature Recalibration:}
The final spectrally attended features are obtained by:
\begin{equation}
F_{\text{attended}} = F_{\text{bdrop}} \odot \text{Reshape}(A_{\text{spec}})
\end{equation}
\subsubsection{\textbf{Spatial Dimension Reduction}}
To prepare features for the transformer backbone, we reduce the spectral dimension through averaging and apply spatial patch embedding:
\begin{equation}
F_{\text{spatial}} = \text{Mean}(F_{\text{attended}}, \text{dim}=2) \in \mathbb{R}^{B \times C_{\text{spec}} \times H_{\text{spec}} \times W_{\text{spec}}}
\end{equation}
\begin{equation}
F_{\text{embed}} = \text{Swish}(\text{BatchNorm2d}(\text{Conv2d}_{4 \times 4, \text{stride}=2}(F_{\text{spatial}})))
\end{equation}
\subsection{\textbf{Hierarchical Transformer Backbone with LoRA Integration}}
The LOLA-SpecViT processes the embedded features through a three-stage hierarchical transformer pipeline. Each stage $i$ consists of a backbone module $\mathcal{B}_i$ followed by a downsampling operation $\mathcal{D}_i$:
\begin{equation}
\mathbf{X}_i = \mathcal{D}_i \left( \mathcal{B}_i \left( \mathbf{X}_{i-1} \right) \right)
\end{equation}
where $\mathbf{X}_0 = F_{\text{embed}}$ is the input to the first stage.
\textbf{Progressive Feature Scaling:}
Each downsampling operation halves the spatial resolution while doubling the channel dimension:
\begin{align}
\mathbf{X}_0 &\in \mathbb{R}^{B \times H_0 \times W_0 \times C_0},\quad
\mathbf{X}_1 \in \mathbb{R}^{B \times H_0/2 \times W_0/2 \times 2C_0}, \\
\mathbf{X}_2 &\in \mathbb{R}^{B \times H_0/4 \times W_0/4 \times 4C_0},\quad
\mathbf{X}_3 \in \mathbb{R}^{B \times H_0/8 \times W_0/8 \times 8C_0}
\end{align}
\subsubsection{\textbf{LoRA-Enhanced Window Attention}}
Each backbone module $\mathcal{B}_i$ integrates Low-Rank Adaptation (LoRA) into the self-attention mechanism for parameter-efficient fine-tuning.
Input features are partitioned into non-overlapping windows of size $M \times M$:
\begin{equation}
\mathbf{W}_j = \text{WindowPartition}(\mathbf{X}_{i-1}, M) \in \mathbb{R}^{B \cdot N_w \times M^2 \times C_i}
\end{equation}
where $N_w = \frac{H_i \cdot W_i}{M^2}$ is the number of windows.
For each window $\mathbf{x}_w \in \mathbb{R}^{M^2 \times C_i}$, queries, keys, and values are computed using LoRA-augmented linear transformations:
\begin{equation}
\begin{aligned}
\mathbf{Q} &= \mathbf{x}_w \mathbf{W}_q + \alpha_q \cdot \mathbf{x}_w \mathbf{A}_q \mathbf{B}_q,\quad
\mathbf{K} = \mathbf{x}_w \mathbf{W}_k + \alpha_k \cdot \mathbf{x}_w \mathbf{A}_k \mathbf{B}_k, \\
\mathbf{V} &= \mathbf{x}_w \mathbf{W}_v + \alpha_v \cdot \mathbf{x}_w \mathbf{A}_v \mathbf{B}_v
\end{aligned}
\label{eq:lora_qkv}
\end{equation}
The matrices $\mathbf{W}_q$, $\mathbf{W}_k$, and $\mathbf{W}_v$ are the frozen, pre-trained weight matrices of shape $\mathbb{R}^{C_i \times d_{\text{head}}}$. In contrast, $\mathbf{A}_q$, $\mathbf{A}_k$, and $\mathbf{A}_v$ are trainable down-projection matrices in $\mathbb{R}^{C_i \times r}$, while $\mathbf{B}_q$, $\mathbf{B}_k$, and $\mathbf{B}_v$ are trainable up-projection matrices in $\mathbb{R}^{r \times d_{\text{head}}}$. Here, the LoRA rank $r$ is chosen such that $r \ll \min(C_i, d_{\text{head}})$, and the scaling factors $\alpha_q$, $\alpha_k$, and $\alpha_v$ are typically set to $\text{lora\_alpha}/r$.
The attention mechanism incorporates relative position bias for better spatial modeling:
\begin{equation}
\text{Attention}(\mathbf{Q}, \mathbf{K}, \mathbf{V}) = \text{Softmax} \left( \frac{\mathbf{Q}\mathbf{K}^T}{\sqrt{d_k}} + \mathbf{B}_{\text{rel}} \right) \mathbf{V}
\end{equation}
where $\mathbf{B}_{\text{rel}} \in \mathbb{R}^{M^2 \times M^2}$ is the relative position bias matrix.
The attention output is processed through a LoRA-enhanced projection:
\begin{equation}
\mathbf{O} = \text{Attention}(\mathbf{Q}, \mathbf{K}, \mathbf{V}) \mathbf{W}_o + \alpha_o \cdot \text{Attention}(\mathbf{Q}, \mathbf{K}, \mathbf{V}) \mathbf{A}_o \mathbf{B}_o
\end{equation}
To stabilize training dynamics and improve gradient flow, we employ \textbf{recharge-based residual connections} with learnable scaling parameters:
\begin{equation}
\mathbf{Y}_{\text{attn}} = \gamma_{\text{attn}} \cdot \mathbf{X} + \beta_{\text{attn}} + \text{DropPath}(\mathbf{O})
\end{equation}
where $\gamma_{\text{attn}}$ and $\beta_{\text{attn}}$ are learnable scalars initialized to 1 and 0 respectively, and DropPath is stochastic depth regularization.
\subsubsection{\textbf{SwiGLU Feed-Forward Network}}
Each transformer block includes a SwiGLU-based feed-forward network for enhanced non-linearity:
\begin{equation}
\text{SwiGLU}(\mathbf{x}) = (\text{Swish}(\mathbf{x}\mathbf{W}_1) \odot (\mathbf{x}\mathbf{W}_2)) \mathbf{W}_3
\end{equation}
where $\mathbf{W}_1, \mathbf{W}_2 \in \mathbb{R}^{C_i \times d_{\text{ffn}}}$ and $\mathbf{W}_3 \in \mathbb{R}^{d_{\text{ffn}} \times C_i}$.
The complete transformer block output is:
\begin{equation}
\mathbf{Y}_{\text{ffn}} = \gamma_{\text{ffn}} \cdot \mathbf{Y}_{\text{attn}} + \beta_{\text{ffn}} + \text{DropPath}(\text{SwiGLU}(\text{LayerNorm}(\mathbf{Y}_{\text{attn}})))
\end{equation}
\subsubsection{\textbf{Downsampling Module}}The downsampling module reduces spatial complexity while preserving semantic information through a sophisticated compression pipeline:
\textbf{Depthwise Convolution and Residual Connection:}
\begin{equation}
\mathbf{X}' = \mathrm{Swish}\bigl(\mathrm{Conv2d}_{3\times 3,\;\mathrm{groups}=C_i}(\mathbf{X})\bigr),\quad
\mathbf{X}'' = \mathbf{X} + \mathbf{X}'.
\end{equation}
\textbf{Squeeze-and-Excitation Attention:}
\begin{equation}
\mathbf{z} = \mathrm{GAP}(\mathbf{X}'') \in \mathbb{R}^{B\times C_i},\;
\mathbf{s} = \sigma\bigl(\mathbf{W}_{\mathrm{se2}}\;\mathrm{Swish}(\mathbf{W}_{\mathrm{se1}}\;\mathbf{z})\bigr),\;
\mathbf{X}''' = \mathbf{s}\odot\mathbf{X}''.
\end{equation}
\textbf{Pointwise Convolution and Spatial Reduction:}
\begin{equation}
\mathbf{Y} = \mathrm{BatchNorm2d}\bigl(\mathrm{Conv2d}_{1\times 1}(\mathbf{X}''')\bigr),\quad
\mathbf{Z} = \mathrm{Conv2d}_{3\times 3,\;\mathrm{stride}=2}(\mathbf{Y})\,. 
\end{equation}
\subsection{\textbf{Classification Head and Output Processing}}
\subsubsection{\textbf{Global Feature Aggregation}}
The final backbone block produces spatial features that are globally aggregated for classification:
\begin{equation}
\mathbf{f}_{\text{global}} = \text{GAP}(\text{LayerNorm}(\mathcal{B}_3(\mathbf{X}_2))) \in \mathbb{R}^{B \times C_{\text{final}}}
\end{equation}
\subsubsection{\textbf{LoRA-Enhanced Classification Head}}
The classification is performed using a LoRA-enhanced linear layer:

\begin{equation}
\hat{\mathbf{y}} = \mathbf{f}_{\text{global}} \mathbf{W}_{\text{cls}} + \alpha_{\text{cls}} \cdot \mathbf{f}_{\text{global}} \mathbf{A}_{\text{cls}} \mathbf{B}_{\text{cls}}
\end{equation}
where $\mathbf{W}_{\text{cls}} \in \mathbb{R}^{C_{\text{final}} \times N_{\text{classes}}}$ is the frozen classification weight matrix, and $\mathbf{A}_{\text{cls}} \in \mathbb{R}^{C_{\text{final}} \times r_{\text{cls}}}$, $\mathbf{B}_{\text{cls}} \in \mathbb{R}^{r_{\text{cls}} \times N_{\text{classes}}}$ are the trainable LoRA adaptation matrices.
\subsubsection{\textbf{Probability Distribution and Prediction}}
The final class probabilities are obtained through softmax normalization:
\begin{equation}
P(c|\mathbf{x}) = \frac{\exp(\hat{y}_c)}{\sum_{j=1}^{N_{\text{classes}}} \exp(\hat{y}_j)}
\end{equation}
The predicted class is determined by:
\begin{equation}
\hat{c} = \arg\max_{c \in \{1,\ldots,N_{\text{classes}}\}} P(c|\mathbf{x})
\end{equation}
LOLA‑SpecViT is a lightweight yet expressive architecture for hyperspectral image classification that markedly reduces trainable parameters via Low‑Rank Adaptation (LoRA) while preserving full model capacity. It fuses 3D spectral convolutions with a hierarchical 2D transformer backbone, capturing both fine spectral signatures and broader spatial patterns. A window‑based attention scheme—with relative position bias and LoRA‑enhanced projections—cuts attention complexity from quadratic to linear, and additional techniques like spectral attention, band‑dropout regularization, and cyclical LoRA scaling bolster robustness and training stability. Its modular design adapts easily to varied spectral resolutions and spatial scales, offering a practical, high‑performance solution for real‑world remote‑sensing tasks.
\section{Experiments}
\label{sec:experiments}
\subsection{\textbf{Dataset}}
In this work, we employ three publicly available hyperspectral datasets WHU-Hi-LongKou, WHU-Hi-HongHu, and Salinas to evaluate our classification framework under diverse agricultural scenarios comprehensively. These datasets differ in spatial resolution, spectral coverage, scene complexity, and number of land-cover classes, providing a robust basis for assessing model generalizability.
\textbf{WHU-Hi-LongKou} dataset was acquired on 17 July 2018 in Longkou Town using a Headwall Nano-Hyperspec sensor mounted on a DJI Matrice 600 Pro UAV. The imagery covers a 550\,×\,400-pixel area at approximately 0.463\,m spatial resolution and spans 270 spectral bands ranging from 400\,nm to 1000\,nm. It depicts a simple agricultural scene containing six main crops: corn, cotton, sesame, broadleaf soybean, narrowleaf soybean, and rice. The provided ground truth defines nine land-cover classes, including the six crops plus water, built-up area, and mixed weeds \cite{RSIDEA2018LongKou}.
\textbf{WHU-Hi-HongHu} dataset was collected on 20 November 2017 in Honghu City using the same Nano-Hyperspec sensor on a DJI Matrice 600 Pro UAV. This image is 940\,×\,475 pixels in size, with 270 spectral bands at ~0.043\,m spatial resolution. The scene represents a complex agricultural testbed with numerous crop types alongside non-crop features such as roads, rooftops, and bare soil. The ground truth comprises 22 land-cover classes across vegetation and infrastructure categories \cite{RSIDEA2017HongHu}.
\textbf{Salinas} scene is an airborne hyperspectral image of agricultural land in Salinas Valley, California, USA. It was acquired by NASA’s AVIRIS instrument at 3.7\,m ground sampling distance. The scene is 512\,×\,217 pixels, with 224 original bands. Describe vegetable fields, vineyards, and bare soils, and is annotated in 16 classes of land cover \cite{CCWintcoSalinas}.
\begin{figure}[htbp]
  \centering
  \includegraphics[width=0.50\textwidth]{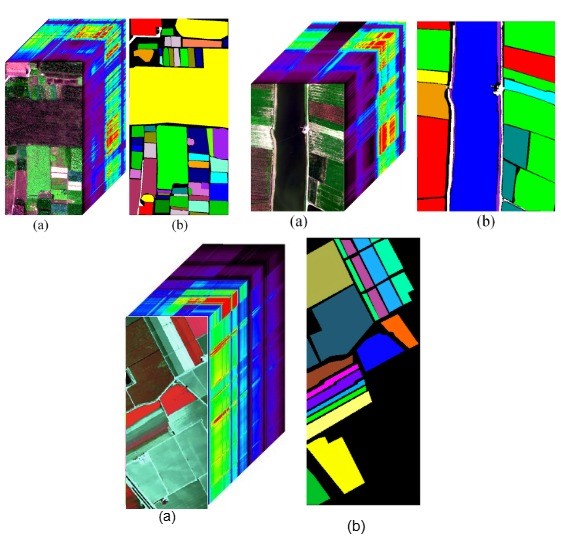}
  \caption{Visualization of hyperspectral data cubes and corresponding ground‐truth classification maps for the WHU‑Hi‑HongH, WHU‑Hi‑Longkou, and Salinas datasets: (a) hyperspectral image cube; (b) ground‑truth map.}
  \label{fig:whu-longkou}
\end{figure}
\subsection{\textbf{Configurations}}
To ensure reproducibility and fair comparison, all experiments were conducted under a consistent hardware and software environment. The implementation was developed using the PyTorch framework and executed on a workstation equipped with an NVIDIA GeForce RTX 3080 GPU and 64\,GB of system RAM.
We adopted the Adam optimizer with an initial learning rate of \(1 \times 10^{-3}\). Models were trained for 100 epochs with a mini-batch size of 64. Each training instance consisted of hyperspectral patches of size 15\,×\,15\,×\,15 (spectral × height × width), selected based on non-overlapping spatial windows.
Our proposed LoLA-SpecViT was configured with three hierarchical stages, with depths of [3, 4, 19], embedding dimension of 96, and head counts of attention of [4, 8, 16]. Each stage employed a windowed self-attention mechanism with a fixed size of 7\,×\,7. For regularization, we applied a drop path rate of 0.2 and integrated LoRA into attention layers using a rank of \(r=16\) and scaling factor \(\alpha=32\).
To improve fine-tuning stability and efficiency, we employed a CLR scheduler designed for LoRA modules. The CLR policy oscillated the internal scaling factor between 0.8 and 1.5 using a `triangular2` schedule with a step size of 100 iterations. This dynamic adjustment encouraged smoother convergence and better generalization across datasets.
\section{Results}
\label{sec:results}
To validate our approach against the fundamental challenges of hyperspectral classification high-dimensional data, limited labels, and complex spectral variability we conduct a series of experiments on three benchmark datasets with only 2\%, 5\%, and 10\% of samples for training. To evaluate the effectiveness of our hierarchical spectral transformer, we compare it against state‐of‐the‐art methods, on the Longkou dataset. First, we compare global performance metrics against these baselines; then, we analyze class-specific results to illustrate how our model overcomes redundancy, captures subtle spectral differences, and mitigates overfitting under severe label scarcity.
\subsection{\textbf{Quantitative Performance Analysis}}
The evaluation of the WHU-Hi-HongHu dataset demonstrates that our proposed framework achieves state-of-the-art performance, attaining an OA of \textbf{99.55\%}, an AA of \textbf{98.85\%}, and a Kappa coefficient of \textbf{99.43\%}. These results outperform the best existing method, DBCT, by 0.41\%, 0.78\%, and 0.52\%, respectively, as shown in Table~\ref{tab:honghu-full}. These gains are particularly notable considering the experiments were conducted using only 10\% of the available labeled data. The robustness of our approach under such a constrained supervision regime highlights the effectiveness of our spectral-processing front-end, which leverages BandDropout and channel-attention modules to emphasize discriminative bands and suppress noise. Furthermore, the integration of LoRA for parameter-efficient adaptation and SwiGLU activations for improved non-linearity contribute to a more expressive and stable learning process, enabling the model to capture long-range spectral-spatial dependencies without overfitting.

Class-level comparisons further substantiate these improvements. The Cotton category achieves 99.93\%, surpassing DBCT by 0.24\%, which illustrates the model’s ability to capture intra-class spectral variations caused by factors such as crop density and soil background. The Rape class, which is frequently misclassified due to spectral similarities with other Brassica species, sees an improvement of 0.36\%, reaching 99.83\%. Similarly, Film Covered Lettuce, a spectrally complex class affected by reflectance distortions due to plastic mulch, achieves 99.79\%, outperforming LSGA by 0.63\%. These results, along with consistent performance across both vegetative and non-vegetative classes, confirm the proposed model’s superior ability to model fine-grained spectral-spatial patterns and its generalization to heterogeneous and ambiguous regions in real-world HSI scenes.

\begin{table}[ht]
  \centering
  \caption{Overall, average, kappa and per‑class accuracies (\%) on the HongHu dataset (10\% training samples) for Ours, HSIMAE~\cite{Ibanez2022}, HybridSN~\cite{Roy2020}, ViT~\cite{Dosovitskiy2021}, MASSFormer~\cite{Sun2024MASSFormer}, DBCT~\cite{Xu2024DBCT} and LSGA~\cite{Ma2023LSGA}.}
  \label{tab:honghu-full}
  \resizebox{0.6\linewidth}{!}{%
    \begin{tabular}{lccccccc}
      \hline
      & Ours    & HSIMAE  & HybridSN & ViT     & MASSFormer      & DBCT              & LSGA               \\
      \hline
      Overall Accuracy (\%)        
        & \textbf{99.55} 
        & 96.24 
        & 96.76 
        & 96.76 
        & 98.14 
        & \underline{99.14} 
        & 99.02 \\
      Average Accuracy (\%)        
        & \textbf{98.85} 
        & 92.99 
        & 92.50 
        & 92.50 
        & 96.25 
        & \underline{98.07} 
        & 97.54 \\
      Kappa Score (\%)             
        & \textbf{99.43} 
        & 96.18 
        & 96.77 
        & 96.77 
        & 97.65 
        & \underline{98.91} 
        & 98.75 \\
      \hline
      Red roof                     
        & \textbf{99.59} 
        & 97.40 
        & 98.46 
        & 97.84 
        & 98.99 
        & 98.27 
        & \underline{99.54} \\
      Road                         
        & 96.71 
        & 88.87 
        & 75.45 
        & 87.22 
        & \underline{97.25} 
        & \textbf{97.44} 
        & 96.82 \\
      Bare soil                    
        & \textbf{99.42} 
        & 95.84 
        & 95.61 
        & 94.98 
        & 98.87 
        & \underline{99.01} 
        & 98.23 \\
      Cotton                       
        & \textbf{99.93} 
        & 98.13 
        & 99.40 
        & 98.48 
        & 99.31 
        & \underline{99.69} 
        & 99.65 \\
      Cotton firewood              
        & \underline{98.96} 
        & 78.73 
        & 88.93 
        & 84.61 
        & 94.76 
        & 98.32 
        & \textbf{99.14} \\
      Rape                         
        & \textbf{99.83} 
        & 98.81 
        & 98.41 
        & 98.60 
        & 99.11 
        & \underline{99.47} 
        & 99.34 \\
      Chinese cabbage              
        & \textbf{99.07} 
        & 94.10 
        & 95.77 
        & 92.50 
        & 97.15 
        & \underline{98.81} 
        & 98.01 \\
      Pakchoi                      
        & \textbf{97.34} 
        & 87.53 
        & 93.72 
        & 87.50 
        & 94.16 
        & \underline{98.38} 
        & 96.59 \\
      Cabbage                      
        & 99.33 
        & 98.14 
        & 97.91 
        & 97.91 
        & \underline{99.77} 
        & 99.71 
        & \textbf{99.82} \\
      Tuber mustard                
        & \underline{99.45} 
        & 94.97 
        & 93.73 
        & 92.90 
        & 98.03 
        & 98.53 
        & \textbf{99.94} \\
      Brassica parachinensis       
        & \textbf{99.12} 
        & 91.95 
        & 91.88 
        & 88.10 
        & 95.53 
        & \underline{98.77} 
        & 97.75 \\
      Brassica chinensis           
        & \textbf{99.39} 
        & 92.80 
        & 92.17 
        & 91.23 
        & 98.15 
        & \underline{98.42} 
        & 96.98 \\
      Small Brassica chinensis     
        & \textbf{99.16} 
        & 92.68 
        & 94.22 
        & 91.70 
        & 94.73 
        & \underline{98.48} 
        & 97.63 \\
      Lactuca sativa               
        & \textbf{99.49} 
        & 95.31 
        & 92.83 
        & 91.74 
        & 98.48 
        & \underline{99.02} 
        & 98.90 \\
      Celtuce                      
        & 96.23 
        & 94.56 
        & 87.59 
        & 94.02 
        & \textbf{98.51} 
        & \underline{98.18} 
        & 97.96 \\
      Film covered lettuce         
        & \textbf{99.79} 
        & 98.14 
        & 97.30 
        & 97.69 
        & 97.03 
        & 98.59 
        & \underline{99.16} \\
      Romaine lettuce              
        & 98.93 
        & 96.94 
        & 92.21 
        & 95.59 
        & 98.86 
        & \textbf{99.47} 
        & \underline{99.32} \\
      Carrot                       
        & \textbf{98.96} 
        & 90.37 
        & 89.72 
        & 93.51 
        & 95.90 
        & 95.16 
        & \underline{96.56} \\
      White radish                 
        & \underline{98.94} 
        & 95.63 
        & 91.73 
        & 91.34 
        & 95.61 
        & 90.11 
        & \textbf{99.51} \\
      Garlic sprout                
        & \textbf{98.06} 
        & 88.67 
        & 89.22 
        & 86.62 
        & 93.85 
        & \underline{99.00} 
        & 97.48 \\
      Broad bean                   
        & 98.16 
        & 85.15 
        & 83.35 
        & 84.24 
        & 87.52 
        & \underline{90.77} 
        & \textbf{91.62} \\
      Tree                         
        & \textbf{98.76} 
        & 91.16 
        & 95.38 
        & 88.90 
        & 88.02 
        & \underline{96.76} 
        & 98.08 \\
      \hline
    \end{tabular}%
  }
\end{table}
Building upon these results, we further assess the generalizability of our framework on the Salinas dataset, which presents distinct spectral and spatial characteristics under semi-arid agricultural conditions. As shown in Table~\ref{tab:dataset1-full}, our model achieves an OA of \textbf{99.51\%}, surpassing the previous state-of-the-art MADANet by 0.34 percentage points. The Average Accuracy also reaches \textbf{99.47\%}, exceeding MADANet’s 99.12\% by 0.35\%, while the Kappa score rises to \textbf{99.46\%}, improving upon SSFTT’s 98.46\% by a full percentage point. These consistent gains, achieved with only 10\% of labeled samples, reaffirm the robustness of our design. The adaptive BandDropout mechanism, in conjunction with spectral attention and LoRA-enhanced transformer blocks, contributes to more effective modeling of discriminative spectral–spatial patterns, while SwiGLU activations facilitate better noise filtering and convergence stability during training.
At the class level, our framework demonstrates clear advantages in distinguishing spectrally overlapping or phenologically variable crop types. For example, the class \textit{Grapes\_untrained} attains a precision of \textbf{99.63\%}, outperforming A2MFE by 1.81 percentage points, confirming the model’s ability to resolve background interference from soil and weeds. In \textit{Lettuce\_romaine\_7wk}, which involves subtle variations in canopy maturity and shadowing, our method achieves perfect classification (\textbf{100.00\%}), exceeding HiT by 0.58 percentage points. Additionally, the class \textit{Corn\_senesced\_green\_weeds}, known for its spectral ambiguity due to aging vegetation, is classified with \textbf{99.90\%} accuracy, outperforming SSFTT’s 99.81\%. These improvements, especially in the most challenging categories, underscore the model’s capacity to maintain fine-grained class discrimination across varied agricultural landscapes.
\begin{table}[h]
  \centering
  \caption{Overall, average, kappa and per‑class precision (\%) on the Salinas dataset (10\% training samples) for Ours, MADANet~\cite{cui2023madanet}, A²MFE~\cite{Yang2021A2MFE}, HybridSN~\cite{Roy2020}, SSFTT~\cite{Sun2022}, HiT~\cite{Yang2022a}, and LANet~\cite{Ding2021}.}

  \label{tab:dataset1-full}
  \resizebox{0.8\linewidth}{!}{%
    \begin{tabular}{lccccccc}
      \hline
      & Ours     & MADANet  & A²MFE    & HybridSN & SSFTT     & HiT       & LANet   \\
      \hline
      Overall Accuracy (\%)          
        & \textbf{99.51} 
        & \underline{99.17} 
        & 98.56 
        & 97.05 
        & 98.61 
        & 97.83 
        & 96.67 \\
      Average Accuracy (\%)          
        & \textbf{99.47} 
        & \underline{99.12} 
        & 98.72 
        & 97.52 
        & 98.97 
        & 98.87 
        & 97.12 \\
      Kappa Score (\%)               
        & \textbf{99.46} 
        & 98.34 
        & 97.35 
        & 96.72 
        & \underline{98.46} 
        & 97.58 
        & 96.39 \\
      \hline
      Brocoli\_green\_weeds\_1       
        & \underline{99.83} 
        & 99.64 
        & 99.46 
        & 99.30 
        & 99.34 
        & \textbf{100.00} 
        & 98.32 \\
      Brocoli\_green\_weeds\_2       
        & 99.94 
        & 98.95 
        & 97.28 
        & \underline{99.96} 
        & \textbf{100.00} 
        & 99.75 
        & 98.64 \\
      Fallow                         
        & 98.82 
        & 98.90 
        & 97.99 
        & 98.16 
        & \underline{99.89} 
        & \textbf{100.00} 
        & 95.09 \\
      Fallow\_rough\_plow            
        & 98.65 
        & 97.58 
        & 95.67 
        & 98.05 
        & \underline{99.34} 
        & \textbf{99.85} 
        & 98.01 \\
      Fallow\_smooth                 
        & 98.92 
        & \underline{99.02} 
        & 98.85 
        & 98.42 
        & \textbf{99.50} 
        & 97.59 
        & 98.73 \\
      Stubble                        
        & \textbf{100.00} 
        & 98.01 
        & 98.31 
        & 98.67 
        & 98.31 
        & \underline{99.92} 
        & 98.86 \\
      Celery                         
        & \underline{99.97} 
        & 99.34 
        & 95.76 
        & 98.96 
        & 99.91 
        & \textbf{100.00} 
        & 97.04 \\
      Grapes\_untrained              
        & \textbf{99.63} 
        & 97.77 
        & \underline{97.82} 
        & 92.70 
        & 97.80 
        & 95.81 
        & 96.49 \\
      Soil\_vinyard\_develop         
        & \underline{99.68} 
        & 98.23 
        & 97.56 
        & 98.53 
        & \textbf{100.00} 
        & \textbf{100.00} 
        & 93.11 \\
      Corn\_senesced\_green\_weeds   
        & \textbf{99.90} 
        & 95.91 
        & 95.23 
        & 95.44 
        & \underline{99.81} 
        & 98.19 
        & 94.27 \\
      Lettuce\_romaine\_4wk          
        & 98.54 
        & 98.86 
        & 97.45 
        & 98.08 
        & \underline{98.77} 
        & \textbf{99.42} 
        & 97.19 \\
      Lettuce\_romaine\_5wk          
        & 99.71 
        & 99.23 
        & 98.34 
        & 99.33 
        & \textbf{99.89} 
        & \underline{99.84} 
        & 98.42 \\
      Lettuce\_romaine\_6wk          
        & \underline{99.39} 
        & \textbf{100.00} 
        & 99.21 
        & 99.02 
        & 97.02 
        & \textbf{100.00} 
        & 98.58 \\
      Lettuce\_romaine\_7wk          
        & \textbf{100.00} 
        & 99.34 
        & 98.12 
        & 98.21 
        & 98.96 
        & \underline{99.42} 
        & 98.07 \\
      Vinyard\_untrained             
        & \textbf{98.78} 
        & \underline{98.49} 
        & 94.55 
        & 92.29 
        & 95.52 
        & 92.53 
        & 92.46 \\
      Vinyard\_vertical\_trellis     
        & \textbf{99.75} 
        & 98.36 
        & 93.23 
        & 80.43 
        & 99.49 
        & \underline{99.66} 
        & 90.11 \\
      \hline
    \end{tabular}%
  }
\end{table}
These results on the Salinas dataset confirm that our method consistently excels across datasets with varying spectral complexity, fine-grained class structure, and geographic diversity. To further assess generalizability, we extend our evaluation to the WHU-Hi-LongKou dataset—a scenario characterized by simpler crop types and relatively homogeneous spectral distributions.
The complete comparison is presented in Table~\ref{tab:longkou-10p-methods}, where our proposed framework again achieves the highest performance across all reported metrics. Specifically, it attains an OA of \textbf{99.91\%}, outperforming the strongest competing model, Cross, by \textbf{0.13\%}. Hybrid-ViT and Hir-Transformer, both relying on transformer-based architectures, follow with OA scores of \textbf{99.75\%} and \textbf{99.68\%}, respectively, falling short by at least \textbf{0.16\%}. Our model also achieves the highest Average Accuracy (AA) of \textbf{99.71\%}, providing a decisive improvement over Cross (\textbf{99.45\%}), Hybrid-ViT (\textbf{99.36\%}), and Hir-Transformer (\textbf{99.14\%}). These gains reflect the advantages of our LoRA-enhanced hierarchical transformer backbone, which captures complex spectral-spatial dependencies while maintaining parameter efficiency. Spatial-spectral baselines such as SSMamba and MorpMamba also yield strong results (OA: \textbf{99.51\%} and \textbf{99.70\%}; AA: \textbf{98.45\%} and \textbf{99.25\%}), yet remain outperformed by our framework. Furthermore, our model achieves the highest Kappa score of \textbf{99.89\%}, ahead of Cross (\textbf{99.78\%}), Hybrid-ViT (\textbf{99.68\%}), and Hir-Transformer (\textbf{99.59\%}). By contrast, the reconstruction-based E-SR-SSIM records substantially lower performance across all metrics, reinforcing the limitations of generative methods under low-label constraints.
\begin{table}[ht]
  \centering
  \caption{Comparison of classification performance on the Longkou dataset with 10\% training samples for Ours, Cross~\cite{bai2024cross}, Hybrid‑ViT~\cite{Arshad2024}, Hir‑Transformer~\cite{Ahmad2024}, SSMamba~\cite{ahmad2025spatial}, MorpMamba~\cite{ahmad2025spatial}, and E‑SR‑SSIM~\cite{hu2023improved}.}

  \label{tab:longkou-10p-methods}
  \resizebox{\linewidth}{!}{%
  \begin{tabular}{lccccccc}
    \hline
    Metrics & Ours   & Cross   & Hybrid-ViT & Hir-Transformer & SSMamba & MorpMamba & E-SR-SSIM \\
    \hline
    Overall Accuracy (\%) 
      & \textbf{99.91} 
      & \underline{99.78} 
      & 99.75 
      & 99.68 
      & 99.51 
      & 99.70 
      & 94.43 \\
    Average Accuracy (\%) 
      & \textbf{99.71} 
      & \underline{99.45} 
      & 99.36 
      & 99.14 
      & 98.45 
      & 99.25 
      & 81.78 \\
    Kappa Score (\%)      
      & \textbf{99.89} 
      & \underline{99.78} 
      & 99.68 
      & 99.59 
      & 99.36 
      & 99.61 
      & 92.63 \\
    \hline
  \end{tabular}}
\end{table}
The per-class performance analysis presented in Table~\ref{tab:longkou-10p-ours} highlights our framework’s exceptional capacity to discriminate among all nine land-cover categories using only 10\% of the available training samples. The model achieves an Overall Accuracy of 99.91\%, supported by consistently near-perfect precision, recall, and F1-scores. These results underscore the effectiveness of our hierarchical spectral-transformer architecture under data-scarce conditions. Spectrally homogeneous classes such as \textit{Rice} and \textit{Water} exemplify this robustness, each attaining precision, recall, and F1-scores of 99.99\% and 99.97\%, respectively. \textit{Broad-leaf soybean} achieves near-ideal metrics with 99.95\% precision and 99.96\% recall, yielding an F1-score of 99.96\%. Similarly, \textit{Corn} records a recall of 99.99\% and an F1-score of 99.98\%, indicating that our model successfully captures the full extent of class-relevant spectral–spatial information even with minimal supervision.
The remaining classes further validate the generalization strength of our approach. \textit{Sesame} achieves perfect precision (100\%) but slightly lower recall (99.41\%), resulting in an F1-score of 99.71\%, suggesting minor confusion at class boundaries. \textit{Narrow-leaf soybean}, which presents significant spectral similarity to related crop types, maintains high reliability with 99.79\% recall and a 99.69\% F1-score. Structural categories such as \textit{Roads and Houses} are also well captured, achieving 99.27\% precision and a 99.20\% F1-score. Notably, the challenging \textit{Mixed Weed} class—composed of heterogeneous vegetation—achieves 99.28\% precision and 99.25\% F1-score, reflecting our model's capability to manage complex, noisy signatures. These consistently strong results across both homogeneous and spectrally diverse classes confirm that our LoRA-enhanced hierarchical transformer is highly effective at resolving class boundaries and maintaining classification fidelity in real-world agricultural mapping scenarios.
\begin{table}[]
  \centering
  \caption{Per-class precision, recall, and F1-score for our method (10\% training samples)}
  \label{tab:longkou-10p-ours}
  \resizebox{0.6\linewidth}{!}{%
  \begin{tabular}{lccc}
    \hline
    Class               & Precision            & Recall               & F1-Score            \\
    \hline
    Corn                
      & 0.9996              
      & \textbf{0.9999}     
      & \underline{0.9998}   \\
    Cotton              
      & 0.9985              
      & \textbf{1.0000}     
      & \underline{0.9993}   \\
    Sesame              
      & \textbf{1.0000}     
      & 0.9941              
      & \underline{0.9971}   \\
    Broad-leaf soybean  
      & \underline{0.9995}  
      & \textbf{0.9996}     
      & \textbf{0.9996}      \\
    Narrow-leaf soybean 
      & 0.9960              
      & \textbf{0.9979}     
      & \underline{0.9969}   \\
    Rice                
      & \underline{\textbf{0.9999}}  
      & \underline{\textbf{0.9999}}  
      & \underline{\textbf{0.9999}}  \\
    Water               
      & \underline{\textbf{0.9997}}  
      & \underline{\textbf{0.9997}}  
      & \underline{\textbf{0.9997}}  \\
    Roads and houses    
      & \textbf{0.9927}     
      & 0.9914              
      & \underline{0.9920}   \\
    Mixed weed          
      & \textbf{0.9928}     
      & 0.9921              
      & \underline{0.9925}   \\
    \hline
    \multicolumn{3}{l}{Overall Accuracy (Ours)} 
      & \textbf{0.9991}     \\
    \hline
  \end{tabular}}
\end{table}
Following the comprehensive 10\% evaluation, we further assess the robustness of our model under extreme data scarcity conditions using only 5\% of the available training samples. As reported in Table~\ref{tab:longkou-5p}, our proposed framework maintains outstanding classification performance, achieving an OA of 99.69\%, an AA of 99.05\%, and a Kappa score of 99.59\%. These results emphasize the strong generalization capacity of our hierarchical spectral-transformer, which leverages localized spectral modeling and parameter-efficient adaptation to retain high discrimination power despite the significant reduction in supervision. Core agricultural classes such as \textit{Corn}, \textit{Cotton}, and \textit{Broad-leaf soybean} continue to exhibit near-perfect performance, with F1-scores of 99.91\%, 99.88\%, and 99.79\%, respectively. Similarly, spectrally consistent categories like \textit{Rice} and \textit{Water} achieve F1-scores of 99.68\% and 99.88\%, reflecting the model’s ability to exploit hierarchical feature abstraction for robust spatial-spectral representation under minimal supervision.
The performance across the remaining categories further validates the resilience of our approach. For instance, \textit{Narrow-leaf soybean} reaches a precision of 99.23\% and an F1-score of 98.39\%, while the structural class \textit{Roads and Houses} retains effective recognition with an F1-score of 97.87\%. The challenging \textit{Mixed Weed} class, characterized by spectral heterogeneity, achieves an F1-score of 98.30\%, highlighting the model’s ability to disentangle fine-grained spectral variability. \textit{Sesame}, the most difficult class in this evaluation, obtains a recall of 99.20\% but a slightly reduced precision of 97.77\%, leading to an F1-score of 98.48\%, which nonetheless remains significantly competitive. Collectively, these results demonstrate that our method sustains exceptional classification fidelity across both homogeneous and complex land-cover categories, even when trained with only 5\% of the data. Such robustness under severe data constraints underscores the advantage of our architecture in real-world remote sensing applications, where labeled samples are often limited.
\begin{table}[]
  \centering
  \caption{Classification metrics on the Longkou dataset with 5\% training samples}
  \label{tab:longkou-5p}
  \resizebox{0.6\linewidth}{!}{%
  \begin{tabular}{lccc}
    \hline
    Class                 & Precision            & Recall               & F1-score            \\
    \hline
    Corn                  
      & \textbf{0.9996}      
      & 0.9987              
      & \underline{0.9991}   \\
    Cotton                
      & 0.9977              
      & \textbf{0.9999}     
      & \underline{0.9988}   \\
    Sesame                
      & 0.9777              
      & \textbf{0.9920}     
      & \underline{0.9848}   \\
    Broad-leaf soybean    
      & 0.9978              
      & \textbf{0.9981}     
      & \underline{0.9979}   \\
    Narrow-leaf soybean   
      & \textbf{0.9923}     
      & 0.9757              
      & \underline{0.9839}   \\
    Rice                  
      & 0.9957              
      & \textbf{0.9978}     
      & \underline{0.9968}   \\
    Water                 
      & 0.9980              
      & \textbf{0.9996}     
      & \underline{0.9988}   \\
    Roads and houses      
      & \textbf{0.9820}     
      & 0.9755              
      & \underline{0.9787}   \\
    Mixed weed            
      & \textbf{0.9888}     
      & 0.9773              
      & \underline{0.9830}   \\
    \hline
    Overall Accuracy (OA) & \multicolumn{2}{c}{–} & \textbf{0.9969}      \\
    Average Accuracy (AA) & \multicolumn{2}{c}{–} & \textbf{0.9905}      \\
    Kappa                 & \multicolumn{2}{c}{–} & \textbf{0.9959}      \\
    \hline
  \end{tabular}}
\end{table}
The performance of our framework under extreme data scarcity—using only $2\%$ of the Longkou dataset for training—is summarized in Table~\ref{tab:longkou-2p}. Despite the minimal supervision, our method achieves the highest per-class accuracy in seven out of nine land-cover categories, demonstrating exceptional discriminative power under severe constraints. Specifically, for Class~1, our model attains an accuracy of $99.22\%$, outperforming GAHT by $3.32\%$; in Class~2, it reaches $99.78\%$, exceeding AMHFN by $4.47\%$; and in Class~4, the accuracy of $98.64\%$ surpasses AMHFN by $2.39\%$. Additionally, Class~5 records $95.94\%$—slightly higher than GAHT ($95.75\%$); Class~7 improves by $0.43\%$ compared to both GAHT and AMHFN; and Class~9 achieves $97.38\%$, a $4.62\%$ margin over AMHFN. Our model also ties AMHFN in Class~8 with $97.15\%$. While AMHFN holds an edge in Class~3 and Class~6 by $7.35\%$ and $0.86\%$ respectively, our approach maintains the best overall balance between precision and recall across the full label space. Notably, HiT consistently trails behind top-tier methods in every class, further underscoring our model's advantage.
At the aggregate level, the superiority of our hierarchical spectral transformer is even more evident. Our model secures the highest OA of $98.66\%$, outperforming AMHFN ($97.07\%$) and GAHT ($96.82\%$) by margins of $1.59\%$ and $1.84\%$, respectively. The Kappa coefficient further reinforces this lead, with our method achieving $98.24\%$, significantly surpassing AMHFN ($96.17\%$) and GAHT ($95.86\%$). While AMHFN reports the highest AA of $96.12\%$, our AA of $95.22\%$ remains highly competitive, with only a $0.90\%$ difference. Importantly, conventional transformer models—including ViT, PiT, and HiT—exhibit considerable performance degradation in this low-data regime, with OA dropping to $91.45\%$, $91.65\%$, and $93.18\%$, respectively, and Kappa values falling to $88.93\%$, $89.20\%$, and $91.20\%$. These pronounced drops highlight the inherent limitations of vanilla transformer architectures when faced with label scarcity. By contrast, our method's consistent top-tier results across both per-class and global metrics set a new benchmark for hyperspectral classification under extremely constrained training scenarios.
\begin{table}[]
  \centering
  \caption{Classification accuracy on the Longkou dataset with 2\% training samples for ViT~\cite{Dosovitskiy2021}, PiT~\cite{heo2021rethinking}, HiT~\cite{Yang2022a}, GAHT~\cite{mei2022hyperspectral}, AMHFN~\cite{AMHFN2023}.}

  \label{tab:longkou-2p}
  \resizebox{0.7\linewidth}{!}{%
  \begin{tabular}{ccccccc}
    \hline
    Class No. & ViT    & PiT    & HiT    & GAHT             & AMHFN            & Ours             \\
    \hline
    1         & 89.72  & 89.36  & 89.83  & \underline{95.90} & 95.85            & \textbf{99.22}   \\
    2         & 63.64  & 87.43  & 90.17  & 94.82            & \underline{95.31} & \textbf{99.78}   \\
    3         & 75.12  & 48.18  & 88.01  & \underline{94.44} & \textbf{96.16}    & 88.81            \\
    4         & 90.14  & 89.62  & 90.61  & 95.64            & \underline{96.25} & \textbf{98.64}   \\
    5         & 78.96  & 76.77  & 91.62  & \underline{95.75} & 94.47            & \textbf{95.94}   \\
    6         & 96.36  & 95.45  & 96.12  & \underline{97.87} & \textbf{98.09}    & \textbf{98.95}   \\
    7         & 97.55  & 97.55  & 97.57  & \underline{99.03} & \underline{99.03} & \textbf{99.46}   \\
    8         & 94.11  & 91.79  & 93.34  & \underline{96.89} & 97.15            & \textbf{97.15}   \\
    9         & 89.72  & 90.59  & 92.33  & 91.98            & \underline{92.76} & \textbf{97.38}   \\
    \hline
    κ (\%)    & 88.93  & 89.20  & 91.20  & 95.86            & \underline{96.17} & \textbf{98.24}   \\
    OA (\%)   & 91.45  & 91.65  & 93.18  & 96.82            & \underline{97.07} & \textbf{98.66}   \\
    AA (\%)   & 86.15  & 85.19  & 92.18  & \underline{95.81} & \textbf{96.12}    & 95.22            \\
    \hline
  \end{tabular}}
\end{table}
\begin{figure}[htbp]
  \centering
  \includegraphics[width=\textwidth]{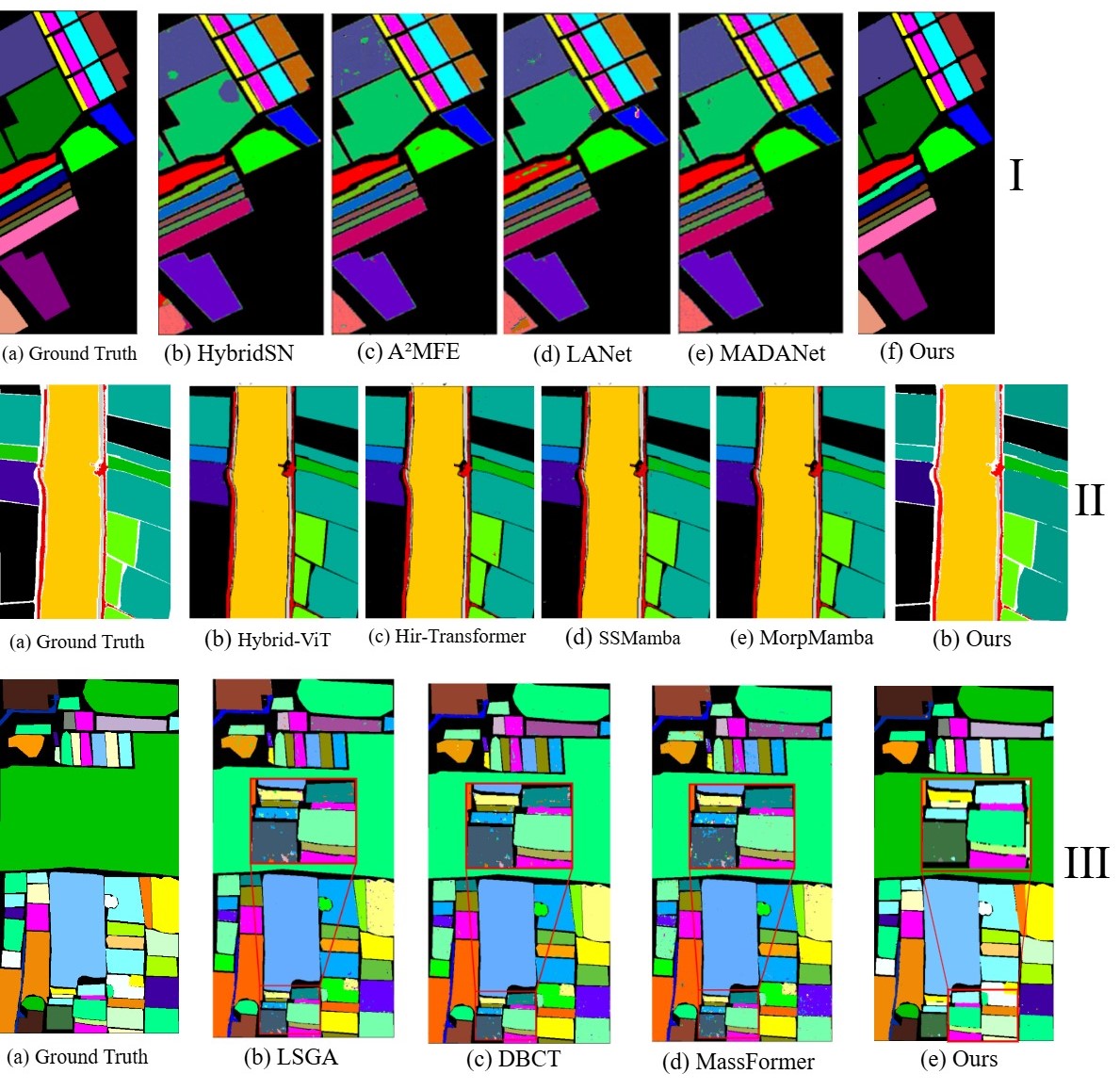}
  \caption{Visual classification results on three HSI benchmarks:  
    \textbf{(I) Salinas dataset}
    \textbf{(II) WHU‑Hi Longkou dataset}  
    \textbf{(III) WHU‑Hi HongHu dataset}}
  \label{fig:Classification map}
\end{figure}

\subsection{\textbf{Visual Classification Analysis}}
Visual evaluation of classification maps offers critical insight into the spatial coherence and semantic integrity of hyperspectral models, particularly in applications where precise boundary delineation and intra-class uniformity are essential. Figure~\ref{fig:Classification map} presents the classification outputs for all three benchmark datasets, revealing the superior visual performance of our proposed framework. Unlike conventional and transformer-based baselines that often suffer from spatial artifacts, boundary blurring, and salt-and-pepper noise, our method consistently produces clean, well-structured classification maps with sharp transitions and minimal misclassification.
This visual superiority is a direct result of our model's hierarchical spectral transformer design, enhanced by two core architectural innovations: \textbf{LoRA} and the \textbf{SwiGLU activation function}. LoRA enables parameter-efficient fine-tuning by updating low-rank projections within the attention mechanism, significantly improving generalisation and suppressing overfitting-induced noise in low-data regimes. SwiGLU introduces a gated nonlinearity that enhances the model’s expressive power and gradient flow, making it particularly effective for capturing subtle spectral transitions and refining object boundaries. Additionally, the inclusion of a CLR scheduler further stabilises training by periodically adjusting the learning rate, thereby facilitating convergence to flatter minima and contributing to smoother and more coherent segmentation outputs.
\subsubsection{\textbf{Salinas Dataset Visual Performance}}

In the Salinas dataset dominated by large, homogeneous crop parcels, salt-and-pepper noise is particularly conspicuous. HybridSN and A$^2$MFE yield fragmented maps and struggle to enforce intra-field uniformity, while LANet and MADANet improve coherence at the cost of boundary sharpness and exhibit spectral bleeding between adjacent plots. By contrast, our LoLA‑SpecViT employs local attention with LoRA adaptation and SwiGLU gating to enforce smooth within‑field predictions and preserve crisp edges. The resulting maps feature sharply defined parcel boundaries, uniform field interiors, and near-complete noise suppression, corroborating the 0.34\% overall accuracy gain over MADANet reported in Table~\ref{tab:dataset1-full}.
\subsubsection{\textbf{Longkou Dataset Visual Performance}}

The Longkou dataset features intricate geometric patterns, with narrow roads and structures interwoven among dense vegetation. Hybrid‑ViT and Hir‑Transformer preserve coarse shapes but frequently mislabel edges and distort slender classes. By leveraging localised attention windows, LoRA, and SwiGLU spectral gating, our LoLA-SpecViT delivers precise boundary adherence and clear semantic separation. In particular, challenging categories such as roads and houses achieve 99.27\% precision and 99.20\% F1 score (Table~\ref{tab:longkou-10p-ours}), reflecting the model’s ability to maintain geometric fidelity and minimize confusion between classes.
\subsubsection{\textbf{HongHu Dataset Visual Performance}}

The HongHu dataset, characterised by high intra-class variability and ambiguous transition zones, poses significant challenges for segmentation models. LSGA’s graph-based local attention, which emphasises narrow spectral affinities, often disrupts spatial continuity and leads to texture fragmentation. DBCT’s use of aggressive dropout and low‑rank convolutional smoothing effectively reduces noise but inadvertently attenuates high-frequency details, causing thin linear structures—such as irrigation channels and hedgerows—to blur. MassFormer’s hybrid local–global attention preserves overall structure yet relies on patch-based self-attention that can misalign oblique boundaries, resulting in distinct stair‑stepped artefacts along class borders.
In contrast, our LoLA‑SpecViT leverages localised attention windows augmented by low‑rank adaptation (LoRA) and SwiGLU spectral gating to effectively model fine spatial–spectral correlations, consistently preserving narrow features, accurately capturing smooth spectral gradients across vegetation–soil interfaces, and maintaining crisp, geometrically faithful boundaries.

\subsubsection{\textbf{Aggregate Visual Performance Across Datasets}}
Taken together, these classification maps confirm the robustness and generalization capacity of our method across diverse landscapes and annotation densities. The clean spatial predictions and structural integrity observed across all three datasets underscore the effectiveness of combining LoRA's regularisation capacity with SwiGLU's non-linear enhancement and CLR's optimization stability. Beyond numerical accuracy, our framework offers interpretable and semantically coherent outputs—an essential quality for practical deployment in remote sensing, environmental monitoring, and precision agriculture, particularly under constrained supervision scenarios (5\%–10\% labeled data).
\subsection{\textbf{Ablation Study}}
To quantify the effect of each architectural component in our framework, we conduct an ablation study on the Longkou dataset, as detailed in Table~\ref{tab:longkou_ablation}. Starting with the \emph{Full Global Context Vision Transformers (GCViT) (4‐stage)} baseline (using GELU activation, without LoRA or CLR), the model attains an OA of \(90.83\%\), AA of \(95.25\%\), and Kappa of \(88.28\%\). Although a four‐stage design is highly expressive, training it from scratch with only \(10\%\) of the labels leads to substantial overfitting and unstable optimization, limiting its generalization performance.

Upon removing one transformer stage (\emph{3‐stage GCViT with only local attention}), OA increases to \(94.05\%\) while AA remains approximately constant at \(95.26\%\), and Kappa improves to \(92.27\%\). This moderate gain indicates that reducing model depth alleviates overfitting and improves gradient flow. However, the truncated architecture with only local attention loses some capacity to capture deep spectral-spatial dependencies; hence, AA does not improve significantly. In other words, the three‐stage variant is less prone to memorizing the small training set but sacrifices fine-grained representation power.

Introducing \emph{LoRA (PEFT)} to the 3‐stage GCViT with local attention (while retaining GELU and no CLR) yields a dramatic jump to OA \(99.91\%\), AA \(99.76\%\), and Kappa \(99.89\%\). This near‐perfect performance demonstrates that parameter-efficient fine-tuning via LoRA enables the model to leverage pre‐trained weights and adapt to the Longkou data distribution with minimal additional parameters. By updating only low-rank attention projections, LoRA effectively regularizes the model, prevents overfitting, and ensures stable convergence even when training data is scarce.

Finally, replacing the GELU activation with \emph{SwiGlu} and enabling the \emph{CLR Scheduler} maintains OA at \(99.91\%\) while AA slightly decreases to \(99.71\%\) and Kappa to \(99.88\%\). The incorporation of CLR allows the learning rate to oscillate periodically, facilitating escaping shallow local minima and accelerating convergence. SwiGlu's gated activation enhances nonlinearity and expressive capacity, particularly for modeling subtle spectral signatures. The minor drop in AA (from \(99.76\%\) to \(99.71\%\) suggests that although SwiGlu and CLR improve optimization stability, the marginal difference can be attributed to small fluctuations in per‐class performance akin to trade‐offs observed when replacing one activation function with another.

\begin{table}[]
\centering
\caption{Ablation Study of our model on the Longkou Dataset}
\label{tab:longkou_ablation}
\resizebox{\linewidth}{!}{%
\begin{tabular}{lccccccccc}
\toprule
\multicolumn{10}{c}{\textbf{Dataset 1: Longkou Dataset}} \\
\midrule
\textbf{Model} & \textbf{4-stage GCViT} & \textbf{3-stage GCViT (Local)} & \textbf{LoRA (PEFT)} & \textbf{CLR Scheduler} & \textbf{GELU} & \textbf{SwiGlu} & \textbf{OA (\%)} & \textbf{AA (\%)} & \textbf{Kappa} \\
\midrule
& \cmark & \xmark & \xmark & \xmark & \cmark & \xmark & 90.83 & 95.25 & 88.28 \\
& \xmark & \cmark & \xmark & \xmark & \cmark & \xmark & 94.05 & 95.26 & 92.27 \\
& \xmark & \cmark & \cmark & \xmark & \cmark & \xmark & \textbf{99.91} & \textbf{99.76} & \textbf{99.89} \\
& \xmark & \cmark & \cmark & \cmark & \xmark & \cmark & \textbf{99.91} & 99.71 & 99.88 \\
\bottomrule
\end{tabular}%
}
\end{table}
\FloatBarrier

To assess the contribution of each architectural component in our framework, we perform an ablation study on the Salinas dataset, as summarised in Table~\ref{tab:salinas_ablation}. The baseline configuration (\emph{Full GCViT (4-stage)} with GELU activation, without LoRA or CLR) yields an OA of \(66.37\%\), AA of \(72.00\%\), and Kappa of \(62.82\%\). This underwhelming performance indicates that the full four‐stage GCViT architecture, when trained from scratch on limited data, suffers from overfitting and fails to capture discriminative spectral‐spatial features properly.

When the fourth stage is removed (\emph{3-stage GCViT with local attention only}), OA slightly increases to \(66.72\%\) and Kappa to \(63.30\%\), but AA decreases to \(70.00\%\). The marginal OA improvement suggests that reducing model depth alleviates overfitting to some extent; however, the drop in AA reveals that the truncated architecture with only local attention loses the capacity to represent fine‐grained variations across certain land‐cover classes, leading to uneven per‐class performance.

Introducing \emph{LoRA} to the three‐stage GCViT with local attention (while retaining GELU and no CLR) dramatically improves all metrics to OA \(99.37\%\), AA \(99.34\%\), and Kappa \(99.42\%\). This substantial leap confirms that parameter‐efficient fine‐tuning via LoRA allows the model to leverage pre‐trained weights while adapting to dataset‐specific distributions with minimal additional parameters. By updating only low‐rank attention matrices, LoRA mitigates overfitting and ensures stable gradient flow, enabling near‐perfect classification even with limited labels.

Finally, replacing GELU with \emph{SwiGlu} and enabling the \emph{CLR Scheduler} further boosts performance to OA \(99.51\%\), AA \(99.46\%\), and Kappa \(99.45\%\). The CLR scheduler's cyclical adjustment of the learning rate helps the optimiser escape shallow local minima and converge more effectively, while SwiGlu's gated activation enhances the network's expressive power for modeling subtle spectral correlations. Consequently, this configuration yields the highest accuracy and inter-class agreement, demonstrating that the combined effect of adaptive learning rates and a more expressive activation function is necessary to achieve robust classification across all Salinas classes under low-label conditions.

\begin{table}[ht]
\centering
\caption{Ablation Study on Salinas Dataset}
\label{tab:salinas_ablation}
\resizebox{\linewidth}{!}{%
\begin{tabular}{lccccccccc}
\toprule
\multicolumn{10}{c}{\textbf{Dataset 2: Salinas Dataset}} \\
\midrule
\textbf{Model} & \textbf{4-stage GCViT} & \textbf{3-stage GCViT (Local)} & \textbf{LoRA (PEFT)} & \textbf{CLR Scheduler} & \textbf{GELU} & \textbf{SwiGlu} & \textbf{OA (\%)} & \textbf{AA (\%)} & \textbf{Kappa} \\
\midrule
& \cmark & \xmark & \xmark & \xmark & \cmark & \xmark & 66.37 & 72.00 & 62.82 \\
& \xmark & \cmark & \xmark & \xmark & \cmark & \xmark & 66.72 & 70.00 & 63.30 \\
& \xmark & \cmark & \cmark & \xmark & \cmark & \xmark & \underline{99.37} & \underline{99.34} & \underline{99.42} \\
& \xmark & \cmark & \cmark & \cmark & \xmark & \cmark & \textbf{99.51} & \textbf{99.46} & \textbf{99.45} \\
\bottomrule
\end{tabular}%
}
\end{table}

To evaluate the individual contributions of each architectural component within our framework, we conduct an ablation study on the HongHu dataset using four experimental configurations, as summarized in Table~\ref{tab:honghu_ablation}. This study isolates the impact of the hierarchical GCViT structure, the introduction of LoRA for parameter‐efficient fine‐tuning (PEFT), the CLR learning rate scheduler, and the substitution of the GELU activation with SwiGLU.

The baseline configuration \emph{Full GCViT (4-stage)} using GELU activation without LoRA or CLR yields significantly underwhelming performance, achieving an OA of only \(56.88\%\), an AA of \(78.33\%\), and Kappa of \(53.56\%\). This result highlights that although the multi‐stage transformer architecture is expressive, it suffers from overfitting and inefficient gradient flow when trained from scratch on limited data without fine‐tuning or learning rate control mechanisms.

When one GCViT stage is removed (\emph{3-stage GCViT with local attention only}), the OA improves to \(66.72\%\). At the same time, AA decreases to \(70.73\%\), and Kappa rises to \(63.30\%\)—this configuration benefits from reduced model complexity, which alleviates overfitting to some extent. However, removing the fourth stage and using only local attention reduces the model's ability to capture long‐range spectral dependencies, leading to uneven per‐class performance (as reflected in the AA drop). This result suggests that deeper transformers may not generalize well in low‐label regimes without appropriate regularization or adaptive training.

Adding LoRA fine‐tuning to the 3-stage GCViT with local attention (without CLR or SwiGLU) leads to a dramatic performance improvement. The OA increases to \(99.30\%\), AA to \(98.17\%\), and Kappa to \(99.11\%\). This clearly illustrates the efficacy of LoRA in enabling parameter‐efficient adaptation without overfitting. By fine‐tuning only low‐rank attention projections, LoRA ensures that the model retains the generalization ability of the pre-trained transformer while adapting to dataset‐specific patterns with minimal overhead.

Finally, replacing GELU activation with SwiGLU and introducing the CLR scheduler yields the highest performance. The model achieves an OA of \(99.54\%\), AA of \(98.84\%\), and Kappa of \(99.42\%\). SwiGLU, a gated activation unit, enhances the expressiveness and non-linearity of the model, especially for spectral sequences with subtle variance. Meanwhile, the CLR scheduler dynamically modulates the learning rate to escape local minima and accelerate convergence, which is particularly beneficial when optimizing deep transformers with sparse labels.

\begin{table}[ht]
\centering
\caption{Ablation Study on Honghu Dataset}
\label{tab:honghu_ablation}
\resizebox{\linewidth}{!}{%
\begin{tabular}{lccccccccc}
\toprule
\multicolumn{10}{c}{\textbf{Dataset 3: Honghu Dataset}} \\
\midrule
\textbf{Model} & \textbf{4-stage GCViT} & \textbf{3-stage GCViT (Local)} & \textbf{LoRA (PEFT)} & \textbf{CLR Scheduler} & \textbf{GELU} & \textbf{SwiGlu} & \textbf{OA (\%)} & \textbf{AA (\%)} & \textbf{Kappa} \\
\midrule
& \cmark & \xmark & \xmark & \xmark & \cmark & \xmark & 56.88 & 78.33 & 53.56 \\
& \xmark & \cmark & \xmark & \xmark & \cmark & \xmark & 66.72 & 70.73 & 63.30 \\
& \xmark & \cmark & \cmark & \xmark & \cmark & \xmark & \underline{99.30} & \underline{98.17} & \underline{99.11} \\
& \xmark & \cmark & \cmark & \cmark & \xmark & \cmark & \textbf{99.54} & \textbf{98.84} & \textbf{99.42} \\
\bottomrule
\end{tabular}%
}
\end{table}
\section{Conclusion}
\label{sec:conclusion}

In this work, we have presented \textbf{LoLA‑SpecViT}, a novel spectral transformer that delivers state‑of‑the‑art hyperspectral classification performance using just 2–10\% of labeled data. By integrating a streamlined 3D spectral front‑end, a hierarchical local‑attention SpecViT backbone, and three core enhancements—low‑rank adaptation (LoRA) to reduce fine‑tuning overhead, SwiGLU gating to enrich spectral–spatial interactions, and a cyclic learning‑rate scheduler to stabilize convergence—our model achieves up to a 0.5\% improvement in overall accuracy, average accuracy, and Cohen’s Kappa over established benchmarks on WHU‑Hi‑LongKou, WHU‑Hi‑HongHu, and Salinas.

Comprehensive ablation analyses validate the individual and combined impact of these innovations, with LoRA playing a critical role in combating overfitting under severe label scarcity. Qualitative assessments further demonstrate that LoLA‑SpecViT produces sharply delineated class boundaries, uniform intra‑class regions, and near‑complete noise suppression, facilitating immediate applicability to downstream tasks such as precision agriculture, change detection, and resource monitoring.

Although our experiments focus on agricultural scenes of VNIR, the architectural principles of LoLASpecViT are broadly applicable across spectral domains and sensor modalities. Future research will explore domain-adaptive pretraining on large HSI corpora, ultraefficient spectral embedding strategies, and prompt‑based tuning techniques to extend model generalization and support real‑time inference on edge platforms.

\bibliographystyle{elsarticle-harv} 
\bibliography{cas-refs}

\appendix
\section{Architecture complexity}
\subsection{Computational Complexity Analysis}
\begin{table}[ht]
\small
\centering
\caption{Enhanced Computational Complexity Analysis}
\label{tab:complexity_improved}
\begin{tabular}{@{}lcc@{}}
\toprule
Component & Time Complexity & Space Complexity \\
\midrule
Spectral Conv3D & $\mathcal{O}(BCHW \cdot \sum_{i} k_i^3)$ & $\mathcal{O}(BdHW)$ \\
Spectral Attention & $\mathcal{O}(BCHW)$ & $\mathcal{O}(BCHW)$ \\
Band Dropout & $\mathcal{O}(BCHW)$ & $\mathcal{O}(BCHW)$ \\
Window Attention & $\mathcal{O}(N \cdot M^2 \cdot d)$ & $\mathcal{O}(M^2 \cdot d)$ \\
LoRA Adaptation & $\mathcal{O}(d \cdot r)$ & $\mathcal{O}(d \cdot r)$ \\
Squeeze-Excitation & $\mathcal{O}(BCHW)$ & $\mathcal{O}(BCHW)$ \\
Overall Model & $\mathcal{O}(B \cdot L \cdot N \cdot M^2 \cdot d)$ & $\mathcal{O}(BdHW)$ \\
\bottomrule
\end{tabular}
\end{table}
where $B$ is batch size, $L$ is number of levels, $N$ is number of tokens, $M$ is window size, $d$ is feature dimension, $r$ is LoRA rank, and $k_i$ represents different kernel sizes in spectral processing.
\subsection{Parameter Efficiency Analysis}
The enhanced parameter efficiency with selective LoRA application:
\begin{equation}
\mathcal{P}_{\text{LoRA}} = \sum_{i \in \mathcal{L}_{\text{LoRA}}} r \cdot (d_{\text{in},i} + d_{\text{out},i})
\end{equation}
where $\mathcal{L}_{\text{LoRA}}$ denotes the set of layers with LoRA adaptation (attention projections and classifier).
The parameter reduction ratio becomes:
\begin{equation}
\rho = \frac{\mathcal{P}_{\text{LoRA}}}{\mathcal{P}_{\text{full}}} = \frac{2r \cdot |\mathcal{L}_{\text{LoRA}}| \cdot d}{d^2 \cdot N_{\text{layers}}} = \frac{2r \cdot |\mathcal{L}_{\text{LoRA}}|}{d \cdot N_{\text{layers}}} \ll 1
\end{equation}
For our configuration with $r=16$, $d=96$, $|\mathcal{L}_{\text{LoRA}}| = 26$ (attention layers + classifier), and $N_{\text{layers}} = 26$, $\rho \approx 0.33$, achieving a 67\% parameter reduction while maintaining full model capacity.
\subsection{Cyclical Learning Rate for LoRA}
The cyclical scaling mechanism for LoRA parameters:

\begin{align}
\text{cycle} &= \left\lfloor 1 + \frac{t}{\text{step\_size\_up} + \text{step\_size\_down}} \right\rfloor \\
x &= t \bmod (\text{step\_size\_up} + \text{step\_size\_down}) \\
\text{scale} &= \begin{cases}
\frac{x}{\text{step\_size\_up}} & \text{if } x < \text{step\_size\_up} \\
1 - \frac{x - \text{step\_size\_up}}{\text{step\_size\_down}} & \text{otherwise}
\end{cases} \\
\gamma_t &= \text{base\_lr} + (\text{max\_lr} - \text{base\_lr}) \cdot \text{scale} \cdot 2^{-(\text{cycle}-1)}
\end{align}
where $t$ is the current iteration, and the triangular2 mode reduces the amplitude by half each cycle.
\section{Algorithms}
\begin{algorithm}[ht]
\small
\caption{PEFT-GCViT Forward Pass}
\label{alg:forward_improved}
\begin{algorithmic}[1]
\Require Hyperspectral input $\mathbf{X} \in \mathbb{R}^{B \times C \times H \times W}$
\Ensure Class predictions $\mathbf{Y} \in \mathbb{R}^{B \times K}$

\State \textbf{Phase 1: Spectral Feature Extraction}
\State $\mathbf{X}' \leftarrow \text{Unsqueeze}(\mathbf{X}, \text{dim}=1)$ \Comment{$\mathbb{R}^{B \times 1 \times C \times H \times W}$}
\State $\mathbf{X}' \leftarrow \text{Conv3D}_{7\times3\times3}(\mathbf{X}') \circ \text{BN3D} \circ \sigma_{\text{swish}}(\mathbf{X}')$ \Comment{1→32 channels}
\State $\mathbf{X}' \leftarrow \text{Conv3D}_{5\times3\times3}(\mathbf{X}') \circ \text{BN3D} \circ \sigma_{\text{swish}}(\mathbf{X}')$ \Comment{32→64 channels}
\State $\mathbf{X}' \leftarrow \text{Conv3D}_{3\times3\times3}(\mathbf{X}') \circ \text{BN3D} \circ \sigma_{\text{swish}}(\mathbf{X}')$ \Comment{64→dim channels}
\State $\mathbf{X}' \leftarrow \text{BandDropout}(\mathbf{X}', p=0.1)$ \Comment{Spectral regularization}
\State $\mathbf{F}_{\text{spec}} \leftarrow \text{SpectralAttention}(\mathbf{X}')$ \Comment{Channel-wise attention}
\State $\mathbf{F}_{\text{spec}} \leftarrow \text{Mean}(\mathbf{F}_{\text{spec}}, \text{dim}=2)$ \Comment{Spectral dimension pooling}

\State \textbf{Phase 2: Patch Embedding \& Positional Encoding}
\State $\mathbf{P} \leftarrow \text{PatchEmbed}(\mathbf{F}_{\text{spec}})$ \Comment{$\mathbb{R}^{B \times \frac{H}{2} \times \frac{W}{2} \times d}$}
\State $\mathbf{P} \leftarrow \mathbf{P} + \mathbf{E}_{\text{pos}}$ \Comment{Add learnable positional encoding}
\State $\mathbf{P} \leftarrow \text{Dropout}(\mathbf{P}, p=0.1)$ \Comment{Positional dropout}

\State \textbf{Phase 3: Hierarchical GCViT Processing}
\For{$\ell = 1$ to $L$} \Comment{$L = 3$ hierarchical levels}
    \For{$b = 1$ to $D_\ell$} \Comment{$D_\ell$ blocks per level $\ell$}
        \State $\mathbf{P} \leftarrow \text{GCViTBlock}_{\ell,b}(\mathbf{P})$
    \EndFor
    \If{$\ell < L$}
        \State $\mathbf{P} \leftarrow \text{ReduceSize}_\ell(\mathbf{P})$ \Comment{Spatial downsampling with SE}
    \EndIf
\EndFor

\State \textbf{Phase 4: Classification Head}
\State $\mathbf{F} \leftarrow \text{LayerNorm}(\text{Flatten}(\mathbf{P}))$
\State $\mathbf{F} \leftarrow \text{GlobalAvgPool}(\mathbf{F})$ \Comment{Global feature aggregation}
\State $\mathbf{Y} \leftarrow \text{LoRALinear}(\mathbf{F})$ \Comment{LoRA-enhanced classifier}
\State \Return $\mathbf{Y}$
\end{algorithmic}
\end{algorithm}

\subsection{Enhanced LoRA with Cyclical Scaling}

\begin{algorithm}[ht]
\small
\caption{Cyclical LoRA-Enhanced Linear Transformation}
\label{alg:lora_cyclical}
\begin{algorithmic}[1]
\Require Input features $\mathbf{x} \in \mathbb{R}^d$, base weights $\mathbf{W} \in \mathbb{R}^{d_{\text{out}} \times d}$
\Require LoRA matrices $\mathbf{A} \in \mathbb{R}^{r \times d}$, $\mathbf{B} \in \mathbb{R}^{d_{\text{out}} \times r}$
\Require Scaling parameters $\alpha, r$, cyclical factor $\gamma_t$
\Require Dropout rate $p_{\text{lora}}$
\Ensure Enhanced output $\mathbf{y} \in \mathbb{R}^{d_{\text{out}}}$

\State $\mathbf{y}_{\text{base}} \leftarrow \mathbf{W}\mathbf{x}$ \Comment{Base transformation}
\State $\mathbf{h} \leftarrow \text{Dropout}(\mathbf{A}\mathbf{x}, p=p_{\text{lora}})$ \Comment{LoRA down-projection with dropout}
\State $\mathbf{y}_{\text{lora}} \leftarrow \mathbf{B}\mathbf{h}$ \Comment{LoRA up-projection}
\State $s_t \leftarrow \frac{\alpha}{r} \cdot \gamma_t$ \Comment{Cyclical scaling factor}
\State $\mathbf{y} \leftarrow \mathbf{y}_{\text{base}} + s_t \cdot \mathbf{y}_{\text{lora}}$ \Comment{Residual connection}
\State \Return $\mathbf{y}$
\end{algorithmic}
\end{algorithm}

\subsection{GCViT Block with PEFT Window Attention}
\begin{algorithm}[ht]
\small
\caption{PEFT-GCViT Block Processing}
\label{alg:gcvit_block_peft}
\begin{algorithmic}[1]
\Require Input tokens $\mathbf{X} \in \mathbb{R}^{B \times H \times W \times d}$
\Require Window size $M$, number of heads $h$, LoRA rank $r$
\Ensure Output tokens $\mathbf{X}' \in \mathbb{R}^{B \times H \times W \times d}$

\State \textbf{Multi-Head Window Attention with LoRA:}
\State $\mathbf{X}_{\text{norm}} \leftarrow \text{LayerNorm}(\mathbf{X})$
\State $\mathbf{W} \leftarrow \text{WindowPartition}(\mathbf{X}_{\text{norm}}, M)$ \Comment{$\mathbb{R}^{N_w \times M^2 \times d}$}
\State $\mathbf{QKV} \leftarrow \text{LoRALinear}(\mathbf{W})$ \Comment{LoRA-enhanced QKV projection}
\State $\mathbf{Q}, \mathbf{K}, \mathbf{V} \leftarrow \text{Split}(\mathbf{QKV}, 3)$ \Comment{Separate Q, K, V matrices}

\For{head $i = 1$ to $h$}
    \State $\mathbf{A}_i \leftarrow \text{Softmax}\left(\frac{\mathbf{Q}_i\mathbf{K}_i^T}{\sqrt{d_k}} + \mathbf{B}_{\text{rel}}\right)$
    \State $\mathbf{O}_i \leftarrow \mathbf{A}_i\mathbf{V}_i$
\EndFor

\State $\mathbf{O} \leftarrow \text{Concat}(\mathbf{O}_1, \ldots, \mathbf{O}_h)$
\State $\mathbf{W}' \leftarrow \text{LoRALinear}(\mathbf{O})$ \Comment{LoRA-enhanced output projection}
\State $\mathbf{X}_{\text{attn}} \leftarrow \text{WindowReverse}(\mathbf{W}', H, W)$
\State $\mathbf{X}_{\text{res1}} \leftarrow \mathbf{X} + \text{DropPath}(\mathbf{X}_{\text{attn}})$

\State \textbf{SwiGLU Feed-Forward Network:}
\State $\mathbf{X}_{\text{norm2}} \leftarrow \text{LayerNorm}(\mathbf{X}_{\text{res1}})$
\State $\mathbf{G}_1, \mathbf{G}_2 \leftarrow \text{Linear}_1(\mathbf{X}_{\text{norm2}}), \text{Linear}_2(\mathbf{X}_{\text{norm2}})$ \Comment{Standard linear layers}
\State $\mathbf{H} \leftarrow \sigma_{\text{swish}}(\mathbf{G}_1) \odot \mathbf{G}_2$ \Comment{SwiGLU gating mechanism}
\State $\mathbf{F} \leftarrow \text{Linear}_3(\mathbf{H})$ \Comment{Final projection}
\State $\mathbf{X}' \leftarrow \mathbf{X}_{\text{res1}} + \text{DropPath}(\mathbf{F})$
\State \Return $\mathbf{X}'$
\end{algorithmic}
\end{algorithm}









\end{document}